\documentclass[letterpaper]{article} 
\usepackage[preprint]{aaai2027}  
\usepackage[hyphens]{url}  
\usepackage{graphicx} 
\usepackage{natbib}  
\usepackage{caption} 
\usepackage{algorithm}

\usepackage{microtype}
\usepackage{newfloat}
\usepackage{listings}
\DeclareCaptionStyle{ruled}{labelfont=normalfont,labelsep=colon,strut=off} 
\floatstyle{ruled}
\newfloat{listing}{tb}{lst}{}
\floatname{listing}{Listing}

\usepackage{booktabs}
\usepackage{titletoc}

\title{Reaching Hidden Populations: Generative Frontier Planning for Peer-Referral Recruitment}
\author{
    Lingkai Kong\textsuperscript{\rm 1}\equalcontrib,
    Hezi Jiang\textsuperscript{\rm 1}\equalcontrib,
    Andrew Ma\textsuperscript{\rm 1},
    Keyu Wang\textsuperscript{\rm 1},\\
    Akseli Kangaslahti\textsuperscript{\rm 1},
    Davin Choo\textsuperscript{\rm 1},
    Milind Tambe\textsuperscript{\rm 1},
    Michael M. Cassell\textsuperscript{\rm 2},\\
    Kayla Stankevitz\textsuperscript{\rm 2},
    Alastair van Heerden\textsuperscript{\rm 3},
    Cheryl Johnson\textsuperscript{\rm 4}
}
\affiliations{
    \textsuperscript{\rm 1}Harvard University\\
    \textsuperscript{\rm 2}FHI 360\\
    \textsuperscript{\rm 3}University of the Witwatersrand\\
    \textsuperscript{\rm 4}World Health Organization
}

\usepackage{amsmath,amssymb,amsfonts,mathtools,amsthm}
\usepackage{algpseudocode}
\usepackage{tikz}
\usetikzlibrary{arrows.meta,positioning, shapes.geometric, fit, backgrounds, calc}

\newcommand{\vY}{\mathbf{Y}}

\newcommand{\E}{\mathbb{E}}
\newcommand{\Prb}{\mathbb{P}}
\newcommand{\N}{\mathbb{N}}
\newcommand{\R}{\mathbb{R}}

\newcommand{\calX}{\mathcal X}
\newcommand{\frontier}{\mathcal F}
\newcommand{\vx}{\mathbf{x}}
\newcommand{\vy}{\mathbf{y}}
\newcommand{\vk}{\mathbf{k}}

\newcommand{\vh}{\mathbf{h}}
\newcommand{\vw}{\mathbf{w}}
\newcommand{\valpha}{\boldsymbol{\alpha}}
\newcommand{\ve}{\mathbf{e}}
\newcommand{\std}[1]{\scriptsize{$\pm$#1}}

\newtheorem{proposition}{Proposition}

\begin{document}

\maketitle

\begin{abstract}
Peer-referral programs help connect populations at high risk for HIV and
other infectious diseases with testing and prevention services when
conventional outreach is ineffective. However, referral vouchers and
incentives are limited, and standard operational guidelines give each
participant the same fixed number of vouchers, ignoring
differences in referral capacity.
Motivated by our collaboration
with public health practitioners, we study how to allocate a fixed voucher
budget adaptively across participants and rounds to maximize total
recruitment.
Prior work
assumes future recruits are i.i.d. from a population-level distribution
and summarizes the recruitment frontier only by its size, ignoring
covariate-dependent recruitment potential. We propose Generative Frontier Planning (GFP), which learns referral
capacity with a censored count model and referrer-conditioned recruit
covariates with a conditional diffusion model. Its structured value
function yields a deterministic Bellman backup shared by fitted value
iteration and planning, while inducing diminishing returns that give
intra-round greedy allocation a \((1-1/e)\)-approximation guarantee for
each fixed round budget. Cross-round selection then determines how much
budget to spend.
We evaluate GFP on two data-calibrated recruitment environments
constructed from de-identified recruiter-recruit records: a public
ICPSR respondent-driven sampling dataset and a dataset provided by
FHI~360, an international global health nonprofit. GFP recruits up to \(21.5\%\) more individuals than the
strongest baseline under the same budget, showing its potential for
resource-limited peer-referral programs.

\end{abstract}

\section{Introduction}

\begin{figure*}[t]
\centering
\resizebox{\linewidth}{!}{%
    \input{fig/problem_summary}%
}%
\caption{
\textbf{GFP for adaptive peer-referral recruitment.}
At round $t$, GFP observes the active frontier
$\frontier_t=\{\vx_1,\ldots,\vx_n\}$ and remaining budget $r_t$, then
allocates $k_i\ge0$ vouchers to each referrer subject to
$\sum_i k_i=s_t\le r_t$ (Steps~1--2). Referrer $i$ has stochastic
capacity $C_i\sim q^\star(\cdot\mid\vx_i)$ and produces
$N_i(k_i)=\min\{k_i,C_i\}$ successful referrals (Step~3). The recruit
covariates are drawn from $G^\star(\cdot\mid\vx_i)$, forming
$\frontier_{t+1}$, while the budget becomes
$r_{t+1}=r_t-s_t$ (Step~4). Unlike fixed-voucher practice and
population-level arrival models~\cite{pan2026adaptive}, GFP adapts
voucher allocations across referrers and models recruits conditionally
on their referrers. The process repeats to maximize the expected
discounted number of recruits. Colors denote covariates, and arrows
denote referrals.
}
    
    \label{fig:recruitment_frontier_pipeline}
\end{figure*}

Many populations at high risk for HIV and other infectious diseases
remain difficult to reach through conventional outreach.
Peer-referral programs address this challenge by recruiting through
trusted social ties, where participants receive vouchers to recruit peers.
In respondent-driven sampling (RDS) \cite{heckathorn1997respondent, goel2009respondent,malekinejad2008using,salganik2004sampling} and related
interventions such as incentivized contact tracing
\cite{munzert2021tracking}, a public health agency begins with a
small set of enrolled participants and gives them vouchers to recruit
peers, who can then recruit others. These programs help estimate disease
prevalence and connect populations that are difficult to reach with
testing, prevention, and other health services.
In collaboration with public health experts from FHI~360, we study how resource-limited programs can maximize outreach.

Operational guidelines for RDS typically allocate the same number of vouchers per participant, regardless of their covariates or recruitment history, e.g., see Annex 6 of \cite{world2013introduction}.
While simple to implement, it does not account for differences in referral ability within the population: some participants can be much more successful in recruiting peers while others may be unable/unwilling to recruit effectively, thus wasting the assigned vouchers.
Moreover, recruitment unfolds sequentially: newly enrolled individuals
become potential referrers and reveal new recruitment opportunities.
An agency can therefore divide a fixed total budget across rounds
and adapt its decisions to the current set of active participants, which
we call the frontier.
Given a limited total budget, the objective is to allocate vouchers across individuals in each frontier to maximize the total number of individuals recruited, as soon as possible.


In this context, \cite{pan2026adaptive} initiated the study of adaptive voucher allocation across multiple rounds.
Under the assumption that each individual's recruitment distribution is drawn i.i.d.\ from a fixed distribution over recruitment distributions, they showed that greedily allocating vouchers within a frontier according to survival probabilities is optimal for a single round.
For the multi-round setting, they approximated the intractable Bellman recurrence with a population-level surrogate based on the expected recruitment distribution.
As a result, the estimated value of a frontier depends only on its size.
In practice, however, recruit characteristics may depend strongly on their referrers: factors such as location, demographics, and risk behavior influence recruitment patterns, with homophily being one well-documented example \cite{goel2009respondent,crawford2018identification}.
Because these characteristics also affect downstream recruitment potential, two frontiers of the same size can have substantially different future values. Consequently, representing a frontier solely by its size can lead to inaccurate value estimates.

Accounting for frontier composition is computationally challenging.
Each allocation induces a different distribution over future frontiers,
and the allocation space is combinatorial. Direct planning would thus
require separate Monte Carlo evaluation of many allocations, making it
impractical even for moderate budgets and frontier sizes.

To address these challenges, we propose Generative Frontier Planning
(GFP). GFP learns referral capacity from censored observations and
referrer-conditioned recruit covariates with a conditional diffusion
model~\cite{ho2020denoising}. It uses a structured value function to evaluate future value without
allocation-specific Monte Carlo sampling and a greedy procedure to
search the allocation space.



We summarize our contributions as follows:
\begin{enumerate}
    \item We formulate peer-referral recruitment as a
    multi-round allocation problem in which referral counts depend on
    individual capacity and voucher allocation, while recruit covariates
    depend on their referrers. We learn these dynamics using a censored
    count model and a conditional diffusion model.

\item We design a structured value function that enables deterministic
next-frontier evaluation through amortized referrer-specific summaries
and yields a monotone diminishing-returns objective, giving greedy
allocation a $(1-1/e)$-approximation guarantee for each fixed budget.

    \item We evaluate GFP in environments calibrated from two
    de-identified recruiter--recruit datasets. GFP outperforms various
    baselines, recruiting up to $21.5\%$ more individuals under the same
    budget.
\end{enumerate}

\section{Problem Formulation}
\label{sec:problem}

We formulate adaptive multi-round peer-referral recruitment as a sequential
planning problem under a total budget \(B\). The goal is to maximize the
expected discounted number of individuals recruited into the program. Each unit
of budget corresponds to one incentivized referral voucher, and a referral is
successful when the referred individual enrolls. We use \(t=0,1,2,\ldots\) to
index recruitment rounds and suppress \(t\) when describing a single round.
Figure~\ref{fig:recruitment_frontier_pipeline} illustrates this.
In the illustration shown in Figure \ref{fig:recruitment_frontier_pipeline}, one should have allocated $k_1 = 2$ to avoid wasting of vouchers and $k_2 = 4$ avoid unused capacity.

\paragraph{State and action.}
In each round, the decision-maker observes a frontier of active participants
\[
    \frontier=\{\vx_1,\ldots,\vx_n\},
\]
where \(\vx_i\in\calX\) contains the observed covariates of participant \(i\),
such as demographic, behavioral, geographic, or network features. The frontier
size \(n=|\frontier|\) may vary across rounds.
The remaining budget \(r\in\N\) is initialized as \(r_0=B\).

An action consists of a round budget $s \in \{0,\ldots,r\}$ and an allocation vector $\vk = (k_1,\ldots,k_n)\in\N^n$ subject to $\sum_{i=1}^n k_i=s$, where $k_i$ is the number of vouchers assigned to participant $i$.
Unused budget $r-s$ can then later be used to allocate vouchers in subsequent rounds.


\paragraph{Referral capacity.}
The referral capacity of participant \(i\) is modeled as
\[
    C_i\sim q^\star(\cdot\mid\vx_i),
\]
where \(q^\star\) is an unknown conditional count distribution and \(C_i\) is
the number of peers whom participant \(i\) is able and willing to recruit.
Given \(k_i\) vouchers, the number of successful referrals is
\[
    N_i(k_i)=\min\{k_i,C_i\}.
\]
Because each successful referral adds one new participant, the immediate reward
is
\[
    R(\frontier,\vk;C)
    =
    \sum_{i=1}^n N_i(k_i),
    \qquad
    C=(C_1,\ldots,C_n).
\]
Vouchers allocated beyond \(C_i\) generate no additional recruits and are
permanently consumed. 


\paragraph{Covariate-dependent generative arrivals.}
In contrast to population-level arrival models in which newly referred
individuals are sampled i.i.d.\ from a fixed population distribution,
we assume that recruit covariates depend on the referrer. Conditional
on \(N_i(k_i)\), the covariates of participant \(i\)'s recruits are
drawn as
\[
    \vx'_{ij}\sim G^\star(\cdot\mid\vx_i),
    \qquad j=1,\ldots,N_i(k_i),
\]
where \(G^\star\) is an unknown referrer-conditioned recruit-covariate
distribution. The next frontier is \[ \frontier' = \bigl\{ \vx'_{ij} : i\in\{1,\ldots,n\},\ j=1,\ldots,N_i(k_i) \bigr\}. \]
Conditioning recruit covariates on the referrer is a central modeling
choice of this work. We let \(G^\star\) depend on the referrer's
covariates and learn it directly from recruiter--recruit records.


\paragraph{Policy and objective.}
A policy \(\pi\) maps each state \((r,\frontier)\) to a feasible action
\((s,\vk)\). After the action, the remaining budget is \(r'=r-s\), and the next
frontier is generated according to \(q^\star\) and \(G^\star\). The value of
\(\pi\) is
\[
    V^\pi(r,\frontier)
    =
    \E_\pi\!\left[
        \sum_{t\ge0}\gamma^t
        R(\frontier_t,\vk_t;C_t)
        \,\bigg|\,
        r_0=r,\ \frontier_0=\frontier
    \right],
\]
where \(\gamma\in(0,1]\). We set
\[
    V^\pi(0,\frontier)=V^\pi(r,\emptyset)=0,
\]
because no further recruitment is possible once the budget is exhausted or the
frontier is empty.

\section{Proposed Method}
\label{sec:method}

We propose Generative Frontier Planning (GFP), a model-based approximate
dynamic programming method for covariate-dependent peer-referral recruitment.
GFP has four components. Section~\ref{subsec:model-fvi} learns how many peers
each participant can recruit and who those recruits are, and identifies the
resulting planning bottlenecks. Section~\ref{subsec:structured-value}
introduces a structured value function for deterministic future-value
evaluation, with a diminishing-returns structure that supports greedy
planning. Section~\ref{subsec:planning} then performs intra-round greedy
allocation for each candidate round budget, followed by cross-round
selection. Finally, Section~\ref{subsec:fitting} learns the structured value
function by reusing the deterministic planner as the backup in fitted
value iteration.

\subsection{Generative Modeling of Referral Dynamics and the Planning Bottleneck}
\label{subsec:model-fvi}

GFP models referral dynamics through two components: a conditional
capacity model $q_\psi(c\mid\vx)$ and a conditional generative model $G_\theta(\vx'\mid\vx)$.

\paragraph{Learning referral capacity from censored observations.}
For each participant, we observe their covariates $\vx_i$, the number of
vouchers $k_i$, and the number of successful referrals
$y_i=\min\{C_i,k_i\}$. If $y_i<k_i$, the participant stopped recruiting
before using all vouchers, so we observe $C_i=y_i$. If $y_i=k_i$, the
vouchers may have limited the observed count, and we only know that
$C_i\ge k_i$.

We model the referral capacity using a Poisson distribution whose rate
depends on the participant covariates:
\[
q_\psi(c\mid\vx)
=
\operatorname{Poisson}
\bigl(c;\lambda_\psi(\vx)\bigr),
\]
where $\lambda_\psi(\vx)$ is a neural network with a softplus output.
We train it by minimizing the censored negative log-likelihood
\begin{equation}
\label{eq:censored-likelihood}
\mathcal L_{\mathrm{cap}}(\psi)
=
-\sum_i
\begin{cases}
\log q_\psi(y_i\mid\vx_i),
& y_i<k_i,\\[3pt]
\log\Prb_{C_i\sim q_\psi(\cdot\mid\vx_i)}
(C_i\ge k_i),
& y_i=k_i.
\end{cases}
\end{equation}
The first case uses an exactly observed capacity, while the second uses
a lower bound on capacity. This prevents participants who use all their
vouchers from being incorrectly treated as having capacity exactly
equal to $k_i$.



\paragraph{Modeling the recruits created by each participant.}
Predicting recruitment counts alone is not enough for multi-round
planning. Two participants may recruit the same number of peers, but
those peers can have very different abilities to continue the
recruitment process. GFP therefore models recruit covariates conditionally
on the referrer using $G_\theta(\vx'\mid\vx)$.
This allows us to distinguish future frontiers that have the same size but different recruitment potential.


We instantiate $G_\theta$ as a conditional diffusion model
\cite{ho2020denoising} that can represent high-dimensional and
multimodal recruit distributions and has been effective in related
public health settings \cite{kangaslahti2026policy}.




\paragraph{The resulting planning bottleneck.}

The two learned models determine both the immediate recruits and the
future frontier generated by an allocation. Given a current frontier
$\frontier$, remaining budget $r$, and allocation $\vk$ with round
budget $s=\sum_i k_i$, the Bellman backup is
\begin{multline}
\label{eq:bellman}
(\mathcal T V_\phi)(r,\frontier)\\
=
\max_{\substack{
0\le s\le r,\;\vk\in\N^n\\
\sum_i k_i=s
}}
\Bigl\{
A_\psi(\frontier,\vk)
+
\gamma\E\!\left[
V_\phi(r-s,\frontier')
\right]
\Bigr\}
\end{multline}

The immediate reward is easy to compute. The $\ell$-th voucher assigned
to participant $i$ produces an additional recruit exactly when
$C_i\ge\ell$. Define its success probability and the resulting expected
immediate reward as
\begin{equation}
\label{eq:immediate-reward}
\begin{aligned}
p_{\psi,i}(\ell)
&=
\Prb_{C_i\sim q_\psi(\cdot\mid\vx_i)}
(C_i\ge\ell),\\
A_\psi(\frontier,\vk)
&=
\sum_{i=1}^n
\sum_{\ell=1}^{k_i}
p_{\psi,i}(\ell).
\end{aligned}
\end{equation}
Thus, the first term $A_\psi(\frontier,\vk)$ in Equation~\eqref{eq:bellman} can be evaluated directly from the capacity model.
However, evaluating the future-value term $\E\!\left[ V_\phi(r-s,\frontier') \right]$ is much harder.

\paragraph{(C1) Allocation-dependent future frontiers.}
Each allocation produces a different distribution over the size and
covariate composition of $\frontier'$. For a general value function,
the expectation
$\E[V_\phi(r-s,\frontier')]$ has no closed form. Direct evaluation would
require separate Monte Carlo rollouts for every candidate allocation.

\paragraph{(C2) Combinatorial allocation space.}
For a fixed round budget $s$, there are
$\binom{s+n-1}{n-1}$ feasible integer allocations across $n$
participants. This number grows rapidly with the budget and frontier
size.

These two challenges reinforce each other: the planner must compare a
large number of allocations, while each comparison may require expensive
simulation of future frontiers.

\subsection{Structured Value Function and Deterministic Future-Value Evaluation}
\label{subsec:structured-value}

To address the two planning bottlenecks, we design the value function
with two goals. First, it should allow the expected value of a random
future frontier to be evaluated without allocation-specific Monte Carlo
sampling, addressing (C1). Second, it should make the combinatorial
allocation problem in (C2) tractable.

\paragraph{Value function parameterization.}
We parameterize the value function as
\begin{equation}
\label{eq:structured-value}
V_\phi(r,\frontier)
=
\sum_{j=1}^d
w_{\phi,j}(r)
\left[
1-
\exp\left(
-\sum_{\vx\in\frontier}
h_{\phi,j}(\vx)
\right)
\right],
\end{equation}
where \(\vh_\phi(\vx)\in\R_+^d\) is the output of a neural network,
\(h_{\phi,j}(\vx)=[\vh_\phi(\vx)]_j\), and
\(w_{\phi,j}(r)\ge0\) is the weight of feature \(j\) under remaining
budget \(r\).

The sum over participants gives a permutation-invariant representation
of a variable-size frontier, while the learned features make the value
sensitive to its covariate composition. The nonnegative features and
weights make \(V_\phi\) monotone: with the same remaining budget, an
additional participant cannot reduce future value because the planner
can assign that participant zero vouchers.

The exponential form serves both design goals. It converts sums over
future recruits into products, enabling deterministic future-value
evaluation. Its saturating shape also induces diminishing returns in the
allocation objective, enabling the greedy optimization in
Section~\ref{subsec:planning}.

For \(r>0\), we set
\[
\vw_\phi(r)=r\cdot\mathrm{softmax}(g_\phi(r)),
\qquad
\vw_\phi(0)=\mathbf 0.
\]
Thus, \(\sum_j w_{\phi,j}(r)=r\), and
\[
V_\phi(0,\frontier)=0,
\qquad
0\le V_\phi(r,\frontier)\le r,
\]
consistent with the fact that each remaining voucher can generate at
most one additional recruit.

\paragraph{Deterministic future-value evaluation.}
The exponential structure reduces the effect of the recruit-covariate
distribution to a referrer-specific summary. For referrer \(\vx\) and
feature \(j\), define
\begin{equation}
\label{eq:laplace-embedding}
\alpha_{\theta,\phi,j}(\vx)
=
\E_{\vy\sim G_\theta(\cdot\mid\vx)}
\left[
\exp\bigl(-h_{\phi,j}(\vy)\bigr)
\right].
\end{equation}
This is the expected multiplicative contribution of one recruit
generated by referrer \(\vx\).

\begin{proposition}[Product-form expected future value]
\label{prop:closed-form}
Fix a frontier \(\frontier=\{\vx_1,\ldots,\vx_n\}\), remaining budget
\(r\), round budget \(s\le r\), and allocation \(\vk\) satisfying
\(\sum_i k_i=s\). Under \(q_\psi\) and \(G_\theta\),
\begin{equation}
\label{eq:closed-form-future}
\begin{aligned}
&\E\!\left[V_\phi(r-s,\frontier')\right]
\\
&\quad =
\sum_{j=1}^d
w_{\phi,j}(r-s)
\left(
1-\prod_{i=1}^n\tau_{ij}(k_i)
\right),
\end{aligned}
\end{equation}
where \(\tau_{ij}(0)=1\) and, for \(k_i\ge1\),
\begin{equation}
\label{eq:tau}
\begin{aligned}
\tau_{ij}(k_i)
&=
\E_{C_i\sim q_\psi(\cdot\mid\vx_i)}
\left[
\alpha_{\theta,\phi,j}(\vx_i)^{\min\{k_i,C_i\}}
\right]
\\
&=
\sum_{c=0}^{k_i-1}
q_\psi(c\mid\vx_i)
\alpha_{\theta,\phi,j}(\vx_i)^c
+
p_{\psi,i}(k_i)
\alpha_{\theta,\phi,j}(\vx_i)^{k_i}.
\end{aligned}
\end{equation}
\end{proposition}

Here, \(\alpha_{\theta,\phi,j}(\vx_i)\) averages over the covariates of
one recruit, \(\tau_{ij}(k_i)\) averages over the number of recruits
from referrer \(i\), and the product across \(i\) combines all
referrers. The second equality in Equation~\eqref{eq:tau} shows that the
count expectation is computed directly from \(q_\psi\), without
sampling. The derivation is given in Appendix~\ref{app:proofs}.

We define the referrer-specific summary vector
\[
\valpha_{\theta,\phi}(\vx)
=
(\alpha_{\theta,\phi,j}(\vx))_{j=1}^d
\in(0,1]^d
\]
and approximate it with an amortization network
\(L_\eta(\vx)\approx\valpha_{\theta,\phi}(\vx)\) with a sigmoid output.
Although \(G_\theta\) remains fixed after training, the target summary
changes as \(h_\phi\) is updated. We therefore periodically refresh \(L_\eta\) using samples from
\(G_\theta(\cdot\mid\vx)\).

For simplicity, in all subsequent planning and training expressions,
\(\tau_{ij}(k_i)\) denotes the plug-in approximation obtained by
substituting \([L_\eta(\vx_i)]_j\) for
\(\alpha_{\theta,\phi,j}(\vx_i)\) in Equation~\eqref{eq:tau}. Thus, planning requires sampling from neither \(G_\theta\) nor
\(q_\psi\).

Combining Equation~\eqref{eq:immediate-reward} with
Proposition~\ref{prop:closed-form} gives
\begin{equation}
\label{eq:surrogate-q}
\begin{aligned}
\widetilde Q_\phi(r,\frontier,s,\vk)
={}&
A_\psi(\frontier,\vk)
\\[-2pt]
&+
\gamma
\sum_{j=1}^d
w_{\phi,j}(r-s)
\left(
1-\prod_{i=1}^n\tau_{ij}(k_i)
\right).
\end{aligned}
\end{equation}
For \(\sum_i k_i=s\), this deterministic approximate \(Q\)-value
combines immediate reward with future recruitment value. It resolves
(C1), while its diminishing-returns structure enables the greedy
solution to (C2) developed next.

\subsection{Intra-Round Greedy Allocation and Cross-Round Selection}
\label{subsec:planning}

Given the deterministic \(Q\)-value
\(\widetilde Q_\phi(r,\frontier,s,\vk)\), GFP plans at two levels.
For a fixed round budget \(s\), it constructs an allocation \(\vk_s\)
across the current frontier, which we call \emph{intra-round allocation}.
It then selects \(s\), trading off budget spent now against budget
reserved for future rounds, which we call \emph{cross-round selection}.

For a fixed \(s\), define
\[
    f_s(\vk)
    \coloneqq
    \widetilde Q_\phi(r,\frontier,s,\vk).
\]
Exact maximization is combinatorial: there are
\(\binom{s+n-1}{n-1}\) feasible allocations satisfying
\(\sum_i k_i=s\), and the future-value term couples participants through
\(\prod_i\tau_{ij}(k_i)\). GFP therefore constructs an allocation
sequentially, one voucher at a time.

To compute marginal gains efficiently, define
\[
    u_j(\vk)
    =
    \prod_{i=1}^n\tau_{ij}(k_i),
    \qquad
    u_j(\mathbf 0)=1.
\]
This quantity tracks the product term in the expected future value.
For a partial allocation \(\vk\), the gain from assigning the next
voucher to participant \(i\) is
\begin{equation}
\label{eq:marginal-gain}
\begin{aligned}
    \Delta_i(\vk;s)
    ={}&
    p_{\psi,i}(k_i+1)
    \\
    &+
    \gamma\sum_{j=1}^d
    w_{\phi,j}(r-s)\,u_j(\vk)
    \left(
        1-
        \frac{\tau_{ij}(k_i+1)}
             {\tau_{ij}(k_i)}
    \right).
\end{aligned}
\end{equation}
The first term is the expected immediate reward from the new voucher,
while the second is its contribution to the expected value of the next
frontier. After assigning the voucher to participant \(i\), the product
terms are updated as
\[
    u_j(\vk+\ve_i)
    =
    u_j(\vk)
    \frac{\tau_{ij}(k_i+1)}
         {\tau_{ij}(k_i)}.
\]

Starting from \(\vk=\mathbf 0\), GFP repeatedly selects
\[
    i^\star
    \in
    \arg\max_i\Delta_i(\vk;s)
\]
and assigns one voucher to \(i^\star\) until all \(s\) vouchers have been
allocated. This produces a candidate allocation \(\vk_s\) for each
round budget \(s\).

After constructing \(\vk_s\) for every feasible round budget, GFP
performs cross-round selection:
\[
    s^\star
    \in
    \arg\max_{0\le s\le r}
    \widetilde Q_\phi
    (r,\frontier,s,\vk_s),
\]
and executes \(\vk_{s^\star}\). Unlike \cite{pan2026adaptive}, where
future value affects only round-budget selection, GFP also uses future
value to determine how vouchers are allocated across participants.
The complete procedure is provided in Algorithm~\ref{alg:gfp} in Appendix~\ref{sec:appendix:alg}.

The structured value function makes \(f_s\) monotone with diminishing
returns, providing the following guarantee.

\begin{proposition}[Diminishing returns and greedy guarantee]
\label{prop:greedy}
Under the structured value function parameterization in
Section~\ref{subsec:structured-value}, \(f_s\) is monotone and satisfies
diminishing returns. Consequently, marginal greedy achieves a
\((1-1/e)\)-approximation to the optimal fixed-budget allocation:
\begin{equation}
\label{eq:greedy-bound}
    f_s(\vk_{\mathrm{greedy}})
    \ge
    (1-1/e)
    \max_{\substack{
        \vk\in\N^n\\
        \sum_i k_i=s
    }}
    f_s(\vk).
\end{equation}
\end{proposition}

\subsection{Fitting the Structured Value Function}
\label{subsec:fitting}

The planner above specifies how to choose an action given a value
function, but the value parameters must still be learned. We fit them
using fitted value iteration, reusing the deterministic \(Q\)-value and
the two-stage planner above to construct approximate Bellman targets.

For a state \((r,\frontier)\) sampled from the training pool and
target parameters \(\bar\phi\), the intra-round greedy procedure
constructs \(\vk_s\) for each round budget \(s\). The resulting target is
\begin{equation}
\label{eq:fvi-target}
\widehat{\mathcal T}V_{\bar\phi}(r,\frontier)
=
\max_{0\le s\le r}
\widetilde Q_{\bar\phi}
(r,\frontier,s,\vk_s).
\end{equation}
Here, \(\widetilde Q_{\bar\phi}\) is Equation~\eqref{eq:surrogate-q}
evaluated using the target value parameters. Its \(\tau_{ij}\) terms
are computed using \(L_\eta\) refreshed to match
\(h_{\bar\phi}\).

We update the current value parameters by minimizing
\begin{equation}
\label{eq:fvi-loss}
\mathcal L_V(\phi)
=
\E_{(r,\frontier)}
\left[
\left(
V_\phi(r,\frontier)
-
\widehat{\mathcal T}V_{\bar\phi}(r,\frontier)
\right)^2
\right].
\end{equation}
We periodically update \(\bar\phi\leftarrow\phi\) and refresh
\(L_\eta\). Thus, value-function fitting and planning use the same
deterministic backup, without sampling from \(G_\theta\) or \(q_\psi\)
during each backup.

\section{Related Work}

\paragraph{Peer-referral recruitment in public health.}
Respondent-driven sampling and related public health interventions use
social ties to reach populations underserved by conventional outreach
\cite{heckathorn1997respondent,salganik2004sampling,goel2009respondent,malekinejad2008using,munzert2021tracking}. Prior work primarily studies sampling, estimation,
and biases induced by network structure and homophily
\cite{goel2009respondent,crawford2018identification}. We instead optimize
how a fixed voucher budget is allocated across participants and rounds.

\paragraph{Adaptive allocation and budgeted MDPs.}
The closest planning formulation assumes i.i.d.\ population-level
arrivals and represents the frontier only by its size
\cite{pan2026adaptive}; GFP instead models referrer-conditioned recruit
covariates and distinguishes frontiers with different compositions.
Budgeted and weakly coupled MDPs address sequential allocation under
limited resources
\cite{boutilier2016budget,carrara2019budgeted,meuleau1998solving,
paruchuri2004towards,dolgov2005computationally}, but generally assume a
fixed set of controlled entities. Branching MDPs allow this set to
change, but focus on extinction or reachability rather than allocation
across referrer-dependent recruitment branches
\cite{etessami2019polynomial}.

\paragraph{Value approximation and greedy allocation.}
Approximate dynamic programming uses features or state aggregation to
construct tractable value approximations
\cite{powell2007approximate,bertsekas2012dynamic,
tsitsiklis1996feature,fadaki2025sequential}. Our parameterization handles variable-size frontiers and factorizes
the expected next-frontier value through referrer-specific summaries,
enabling deterministic evaluation. The per-round planner builds on the classical
$(1-1/e)$ guarantee for monotone objectives with diminishing returns
\cite{nemhauser1978analysis}. Its guarantee applies to each fixed-budget
round, while Bellman planning connects decisions across rounds, rather
than relying on adaptive submodularity
\cite{golovin2011adaptive}.

\begin{table*}[t]
\centering
\caption{
Cumulative rewards (mean $\pm$ standard error), with the highest mean
in each column shown in bold. The shifted setting uses a separately
trained covariate model and $\varepsilon_c = 0.2$. We use $B = 200$,
$|F_0| = 10$, $\gamma = 0.99$, and evaluate over 50 episodes for ICPSR
and 30 episodes for FHI~360.
}
\label{tab:main-paper-gamma-099}

\resizebox{\textwidth}{!}{%
\begin{tabular}{@{}lcccc@{\hspace{10pt}}cccc@{}}
\toprule
& \multicolumn{4}{c}{ICPSR}
& \multicolumn{4}{c}{FHI~360} \\
\cmidrule(lr){2-5}
\cmidrule(lr){6-9}

& \multicolumn{2}{c}{Discounted}
& \multicolumn{2}{c}{Undiscounted}
& \multicolumn{2}{c}{Discounted}
& \multicolumn{2}{c}{Undiscounted} \\
\cmidrule(lr){2-3}
\cmidrule(lr){4-5}
\cmidrule(lr){6-7}
\cmidrule(lr){8-9}

Method
& Clean & Shift
& Clean & Shift
& Clean & Shift
& Clean & Shift \\
\midrule

Random
& 65.0 \std{ 4.0}
& 72.2 \std{ 4.2}
& 65.9 \std{ 4.0}
& 73.2 \std{ 4.3}
& 31.2 \std{ 2.4}
& 53.0 \std{ 3.2}
& 31.6 \std{ 2.5}
& 53.7 \std{ 3.3} \\

Constant-2
& 147.4 \std{ 1.4}
& 131.5 \std{ 1.6}
& 151.4 \std{ 1.4}
& 135.5 \std{ 1.5}
& 79.9 \std{ 2.9}
& 78.6 \std{ 4.3}
& 84.7 \std{ 3.1}
& 82.2 \std{ 4.6} \\

Constant-3
& 131.9 \std{ 2.3}
& 118.2 \std{ 1.7}
& 134.3 \std{ 2.3}
& 120.1 \std{ 1.7}
& 63.3 \std{ 1.7}
& 71.1 \std{ 1.1}
& 65.2 \std{ 1.7}
& 72.9 \std{ 1.1} \\

Constant-5
& 103.3 \std{ 2.2}
& 96.7 \std{ 2.4}
& 104.5 \std{ 2.2}
& 97.7 \std{ 2.5}
& 41.2 \std{ 1.1}
& 51.8 \std{ 1.6}
& 41.8 \std{ 1.1}
& 52.4 \std{ 1.6} \\

Constant-10
& 56.3 \std{ 2.1}
& 61.3 \std{ 2.4}
& 56.7 \std{ 2.2}
& 61.6 \std{ 2.4}
& 22.8 \std{ 0.9}
& 37.5 \std{ 1.7}
& 22.9 \std{ 0.9}
& 37.6 \std{ 1.7} \\

Budget DQN
& 145.2 \std{ 1.5}
& 137.5 \std{ 2.0}
& 147.1 \std{ 1.6}
& 139.4 \std{ 2.1}
& 97.1 \std{ 4.4}
& 97.1 \std{ 4.7}
& 101.0 \std{ 4.5}
& 100.6 \std{ 4.8} \\

Factorized RL
& 115.3 \std{ 3.9}
& 103.6 \std{ 4.7}
& 118.7 \std{ 4.0}
& 107.0 \std{ 4.9}
& 29.6 \std{ 2.3}
& 46.7 \std{ 3.5}
& 30.8 \std{ 2.4}
& 48.5 \std{ 3.7} \\

IID Pop DP
& 188.7 \std{ 0.7}
& 166.6 \std{ 1.8}
& 192.0 \std{ 0.6}
& 169.6 \std{ 1.8}
& 94.7 \std{ 4.6}
& 94.5 \std{ 3.2}
& 97.8 \std{ 4.6}
& 97.7 \std{ 3.2} \\

\midrule
\textbf{GFP (ours)}
& \textbf{193.3} \std{ 0.3}
& \textbf{174.0} \std{ 1.2}
& \textbf{198.6} \std{ 0.2}
& \textbf{179.4} \std{ 1.2}
& \textbf{117.4} \std{ 5.1}
& \textbf{107.7} \std{ 3.6}
& \textbf{122.7} \std{ 5.2}
& \textbf{112.8} \std{ 3.7} \\
\bottomrule
\end{tabular}%
}
\end{table*}

\section{Experiments}

We evaluate GFP in two recruitment environments based on real world peer
referral datasets. We study three questions: whether GFP improves recruitment
under clean dynamics, how well it handles changes in the test environment,
and when planning with individual covariates is most useful.
Full experimental details (including training hyperparameters) are provided in Appendix~\ref{app:environment-details}.

\subsection{Recruitment Environments}
\label{subsec:env}

We construct two data-calibrated peer-recruitment environments from
de-identified recruiter--recruit records for HIV screening: the public
ICPSR respondent-driven sampling dataset \cite{morris2011hiv} and a
dataset provided by FHI~360, an international public health nonprofit. In both datasets, nodes represent individuals
and directed edges represent observed referrals. We fit the recruitment-count and recruit-covariate models from individual covariates, node
out-degrees, and observed referral pairs using the same training
procedure.

Both datasets contain $K=17$ categorical covariate fields. After one-hot
encoding, ICPSR contains $1{,}386$ nodes and $6{,}379$ directed edges,
with $p=72$ covariates per individual. For FHI~360, we remove isolated nodes
and fields with more than $45\%$ missing values, resulting in $9{,}234$
nodes, $7{,}950$ directed edges, and $p=101$ covariates per individual.
The FHI~360 environment is substantially sparser and contains more individuals with little
or no observed recruitment capacity. Additional details are provided in
Appendix~\ref{app:environment-details}.




\subsection{Baselines}
\label{subsec:baselines}

We compare GFP with five types of baselines.

\paragraph{Random.}
Uniformly sample round budget \(s_t\) from \(\{0,\ldots,b_t\}\) and allocate it across the frontier uniformly.

\paragraph{Constant \(k\).}
This policy assigns $k\in\{2,3,5,10\}$ vouchers to frontier individuals in a fixed order until the budget is exhausted. It does not use covariates or learned models.

\paragraph{Budget DQN.}
Budget DQN learns the round budget with a Q-network and allocates the chosen
budget using a fixed greedy rule based on predicted recruitment counts. It
learns when to spend but not within-round allocation.

\paragraph{Factorized RL.}
Factorized RL decomposes the decision into round budget, number of active
recruiters, and recruiter scores. A DeepSets~\cite{zaheer2017deep} encoder represents the frontier,
and the selected recruiters receive the budget through softmax scores with
integer rounding.

\paragraph{IID Population DP.}
The method of \cite{pan2026adaptive} that performs backward induction over remaining budget and frontier size, using a marginal Poisson transition model.
It models future frontier size but not covariate composition.

\subsection{Evaluation Protocol}
\label{subsec:setup}

\paragraph{Episodes.}
Each episode starts with $|\frontier_0|=10$, a total budget $B=200$, and a
maximum of $50$ rounds. An episode ends when the budget is exhausted, the
frontier is empty, or the horizon is reached. We evaluate
$\gamma\in\{0.9,0.99,1.0\}$. The main text reports results for
$\gamma=0.99$, while results for the other discount factors are provided
in Appendix~\ref{sect:appendix:additional-results}.
$\gamma$ is fixed for all methods during training or planning and evaluation.

\paragraph{Metrics.}
Our main metric is the total number of recruits, $\sum_t R_t$, which measures
the reach achieved with a fixed budget. We also report discounted cumulative
reward, $\sum_t \gamma^t R_t$, which gives more weight to recruits found in
earlier rounds.

\begin{table}[t]
\centering
\small
\caption{
Robustness of total recruitment to count contamination at
\(\gamma=0.99\) on ICPSR. The shifted covariate model is fixed within each
environment while \(\epsilon_c\) varies. Results are mean $\pm$ standard
error, with the best mean in bold.
}
\label{tab:noise-sensitivity}
\resizebox{\columnwidth}{!}{
\begin{tabular}{lcccc}
\toprule
Method
& $\epsilon_c{=}0.1$ & $\epsilon_c{=}0.2$ & $\epsilon_c{=}0.3$ & $\epsilon_c{=}0.4$ \\
\midrule
Random
& 73.4\std{4.8} 
& 73.2\std{4.3} 
& 74.9\std{3.7} 
& 82.5\std{3.4} \\
Constant-2
& 146.8\std{1.3} 
& 135.5\std{1.5} 
& 132.1\std{1.4} 
& 125.1\std{1.9} \\
Constant-3
& 131.3\std{1.8} 
& 120.1\std{1.7} 
& 118.1\std{1.8} 
& 110.8\std{2.6} \\
Constant-5
& 99.9\std{2.2} 
& 97.7\std{2.5} 
& 98.8\std{1.8} 
& 94.0\std{1.7} \\
Constant-10
& 59.3\std{2.2} 
& 61.6\std{2.4} 
& 68.7\std{2.0} 
& 70.9\std{2.3} \\
Budget-DQN
& 142.2\std{2.0} 
& 139.4\std{2.1} 
& 133.4\std{2.3} 
& 130.0\std{2.1} \\
Factorized RL
& 159.2\std{3.7} 
& 107.0\std{4.9} 
& 110.4\std{4.8} 
& 111.8\std{3.8} \\
IID-Pop. DP
& 182.7\std{1.1} 
& 169.6\std{1.8} 
& 159.8\std{3.5} 
& 153.8\std{3.3} \\
GFP (ours)
& \textbf{188.1}\std{1.0} 
& \textbf{179.4}\std{1.2} 
& \textbf{163.1}\std{3.7} 
& \textbf{162.2}\std{1.8} \\
\bottomrule
\end{tabular}
}
\end{table}

\paragraph{Test-time distribution shifts.}
In the clean setting, learning and evaluation use the same fitted count
and recruit-covariate models. To assess robustness, we evaluate the
policies without adaptation under shifted recruitment dynamics. We
replace the recruit-covariate model with a separate conditional diffusion
model trained on held-out data and perturb recruitment counts as follows.
For an individual receiving \(k_i\) vouchers, we set
\(m_i^{\mathrm{shift}}=m_i^{\mathrm{clean}}\) with probability
\(1-\epsilon_c\); otherwise, we draw
$m_i^{\mathrm{shift}}
\sim \operatorname{Unif}\{0,\ldots,k_i\}.$
This mechanism preserves
\(0\leq m_i^{\mathrm{shift}}\leq k_i\). Together, the held-out covariate
model and count contamination change both recruit covariates and
recruitment counts.


\subsection{Main Results}
\label{subsec:results}

Table~\ref{tab:main-paper-gamma-099} reports the main results at
\(\gamma=0.99\). Results for \(\gamma\in\{0.9,1.0\}\) are provided in
Appendix~\ref{sect:appendix:additional-results}. Our main findings are
summarized below.

\paragraph{GFP achieves the strongest overall performance.}
Across the two environments, two test conditions, and two metrics, GFP
achieves the highest mean in all eight comparisons. Under clean dynamics,
GFP improves total recruitment over the strongest baseline by \(6.6\) in
ICPSR and \(21.7\) in FHI~360; under shifted dynamics, the gains are \(9.8\)
and \(12.2\), respectively. GFP also improves discounted reward
consistently, with gains of \(4.6/20.3\) under clean dynamics and
\(7.4/10.6\) under shifted dynamics for ICPSR/FHI~360. These results show
that GFP remains effective across different network structures and under
changes in both recruitment counts and recruit covariates.

\paragraph{GFP benefits from modeling frontier composition.}
The advantage of GFP is especially pronounced in FHI~360, which is sparser
and contains more individuals with limited recruitment capacity. Compared
with IID Population DP, GFP recruits \(24.9\) more individuals in FHI~360
under clean dynamics, compared with \(6.6\) in ICPSR; under shifted
dynamics, the gains are \(15.1\) and \(9.8\), respectively. Since IID
Population DP models only future frontier size, while GFP models its
covariate composition, these results demonstrate the value of
covariate-aware planning when frontier composition affects future
recruitment potential.

\subsection{Robustness to Count Shifts}
\label{subsec:noise}

We evaluate GFP’s robustness to shifts in recruitment counts by holding the shifted covariate model fixed while varying the count contamination level, \(\epsilon_c\).

Table~\ref{tab:noise-sensitivity} reports total
recruitment at \(\gamma=0.99\) on ICPSR. As we can see, GFP achieves the best performance at every contamination level. Compared with IID Population DP, GFP improves recruitment
by \(3.3\)--\(9.8\). As
$\epsilon_c$ increases, recruitment counts depend less on individual
characteristics, making strong recruiters more difficult to identify.
Nevertheless, GFP remains the best-performing method across all
settings.

\subsection{Ablation of Referrer Conditioning}
\label{subsec:parent-ablation}

To evaluate the effect of referrer--recruit dependence, we replace the
conditional diffusion model
$p(\mathbf{x}_{\mathrm{recruit}}
\mid \mathbf{x}_{\mathrm{referrer}})
$
with an unconditional model
$p(\mathbf{x}_{\mathrm{recruit}}),$
while keeping the remaining architecture unchanged.

\begin{figure}[t]
  \centering
  \includegraphics[width=\columnwidth]{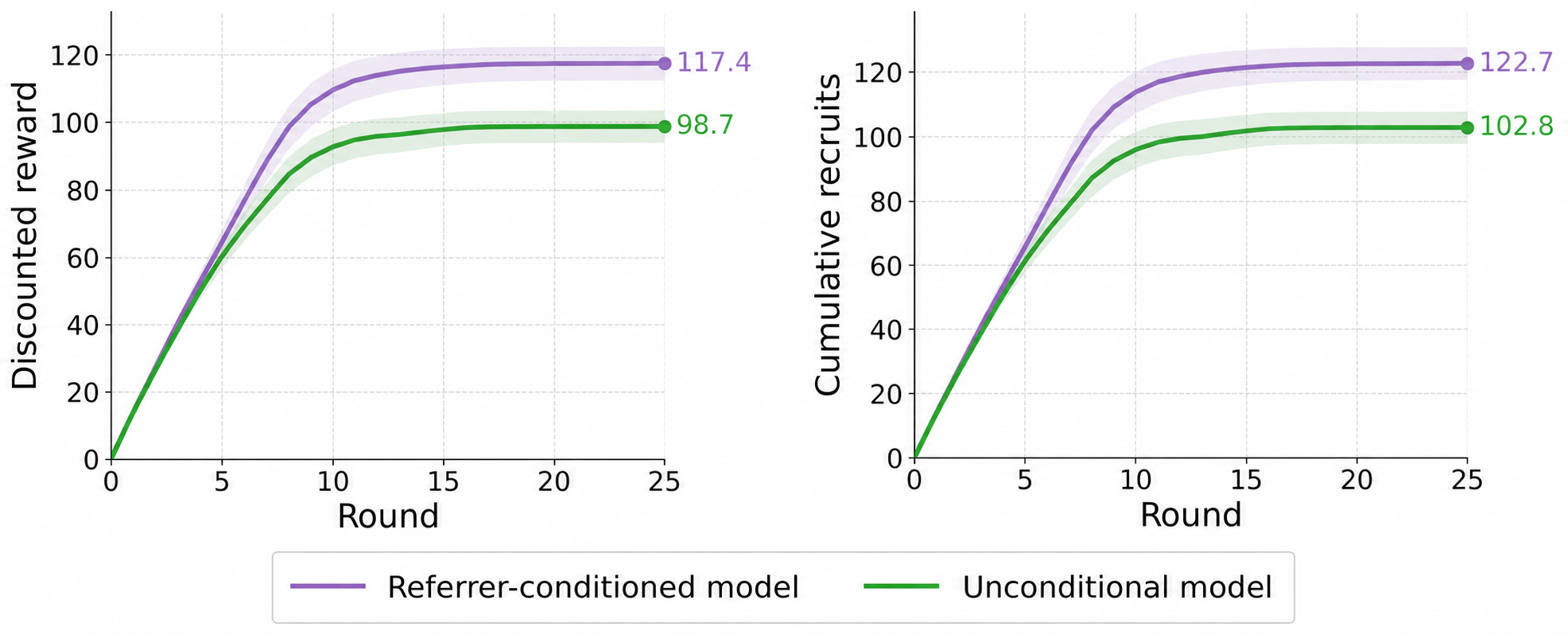}
  \caption{
Cumulative discounted reward (left) and cumulative recruits (right) for
GFP with referrer-conditioned recruit modeling and its unconditional
ablation at \(\gamma=0.99\). Lines show means over 30 episodes with \(\pm1\) SE bands.
  }
  \label{fig:parent-conditioning-ablation}
\end{figure}

Figure~\ref{fig:parent-conditioning-ablation}
shows the corresponding allocation curves on FHI~360 under clean dynamics with \(\gamma=0.99\). As we can see, the conditional model
achieves \(122.7\pm5.2\) cumulative recruits and
\(117.4\pm5.1\) discounted reward, compared with
\(102.8\pm5.1\) and \(98.7\pm4.9\) for the unconditional model. This
corresponds to improvements of \(19.3\%\) and \(18.9\%\), demonstrating that referrer--recruit dependence improves
future-value estimation and downstream allocation.




\section{Conclusion and Path Forward}

We introduced Generative Frontier Planning (GFP) for adaptive
peer-referral recruitment. GFP learns referral dynamics by combining
a censored capacity model with a conditional diffusion model that
captures referrer-dependent recruit covariates. These learned dynamics
enable GFP to distinguish frontier compositions with different
downstream recruitment potential. Its structured value function
provides deterministic expected-value evaluation and supports a
per-round greedy planner with a $(1-1/e)$-approximation guarantee.
Across two data-calibrated environments, GFP consistently outperforms
existing allocation methods, demonstrating the value of generative
modeling for covariate-aware planning under limited recruitment
resources.

Looking forward, we plan to continue collaborating with FHI~360 and our other public health expert co-authors to adapt GFP to operational needs, incorporate practitioner feedback, and evaluate the system under real-world constraints. These efforts will
provide a path toward the eventual deployment of GFP as a decision-support
tool in peer-referral programs.

\section{Ethical considerations}
This work develops decision-support methods for adaptive peer-referral
recruitment in public health settings. Since prioritizing individuals
based on estimated recruitment potential may reflect or amplify existing
inequalities in network access and participation, GFP should be deployed
with appropriate domain oversight and fairness safeguards rather than as
an automated policy prescription. The ICPSR dataset \cite{morris2011hiv}
is a publicly available de-identified dataset accessed under ICPSR's
terms of use and does not require IRB approval. The FHI~360 dataset is used
under an IRB-approved data sharing agreement with FHI~360.


\bibliography{refs}

\newpage
\newpage
\appendix

\newpage
\appendix

\section{Algorithm}
\label{sec:appendix:alg}

Algorithm~\ref{alg:gfp} summarizes the two-stage planner used both for
action selection and for constructing fitted-value-iteration targets.
For each candidate final round budget \(s\), the continuation budget
\(r-s\) is held fixed while GFP constructs an intra-round allocation
one voucher at a time. The planner maintains the product terms
\(u_j(\vk)=\prod_i\tau_{ij}(k_i)\), compares the resulting allocations
across candidate round budgets, and returns the allocation with the
largest deterministic approximate Q-value.

The algorithm below is written using \(w_\phi\) and a summary network
\(L_\eta\) refreshed to match \(h_\phi\). When constructing fitted-value-
iteration targets, \(w_\phi\) and \(h_\phi\) are replaced by
\(w_{\bar\phi}\) and \(h_{\bar\phi}\), respectively, and \(L_\eta\) is
refreshed accordingly.

\begin{algorithm}[H]
\caption{Generative Frontier Planning}
\label{alg:gfp}
\begin{algorithmic}[1]
\Require Frontier \(\frontier=\{\vx_1,\ldots,\vx_n\}\), remaining
budget \(r\), count model \(q_\psi\), refreshed referrer-summary
network \(L_\eta\), value weights \(\vw_\phi\), discount factor
\(\gamma\)
\For{\(s=0,1,\ldots,r\)}
    \State \(\vk\gets\mathbf 0\)
    \State \(u_j\gets1\) for \(j=1,\ldots,d\)
    \For{\(m=1,\ldots,s\)}
        \For{\(i=1,\ldots,n\)}
            \For{\(j=1,\ldots,d\)}
                \State Compute \(\tau_{ij}(k_i)\) and
                \(\tau_{ij}(k_i+1)\) using Equation~\eqref{eq:tau}
                with \([L_\eta(\vx_i)]_j\)
            \EndFor
            \State
            \[
            \Delta_i\gets
            p_{\psi,i}(k_i+1)
            +
            \gamma\sum_{j=1}^d
            w_{\phi,j}(r-s)u_j
            \left(
                1-
                \frac{\tau_{ij}(k_i+1)}
                     {\tau_{ij}(k_i)}
            \right)
            \]
        \EndFor
        \State \(i^\star\gets\arg\max_i\Delta_i\)
        \For{\(j=1,\ldots,d\)}
            \State
            \[
            u_j\gets
            u_j
            \frac{\tau_{i^\star j}(k_{i^\star}+1)}
                 {\tau_{i^\star j}(k_{i^\star})}
            \]
        \EndFor
        \State \(k_{i^\star}\gets k_{i^\star}+1\)
    \EndFor
    \State \(\vk_s\gets\vk\)
    \State
    \[
    \widetilde Q_s\gets
    A_\psi(\frontier,\vk_s)
    +
    \gamma\sum_{j=1}^d
    w_{\phi,j}(r-s)(1-u_j)
    \]
\EndFor
\State \(s^\star\gets
\arg\max_{0\le s\le r}\widetilde Q_s\)
\State \Return \(s^\star\) and \(\vk_{s^\star}\)
\end{algorithmic}
\end{algorithm}

\section{Proofs}
\label{app:proofs}

Throughout the proofs, define the survival probability of the count
model by
\[
p_{\psi,i}(\ell)
=
\Prb_{C\sim q_\psi(\cdot\mid\vx_i)}
(C\ge \ell),
\]
and define the exact referrer-specific summary
\[
\alpha_{ij}
=
\alpha_{\theta,\phi,j}(\vx_i)
=
\E_{\vy\sim G_\theta(\cdot\mid\vx_i)}
\left[
    \exp\bigl(-h_{\phi,j}(\vy)\bigr)
\right].
\]
Since \(h_{\phi,j}(\vy)\ge0\), we have
\(\alpha_{ij}\in(0,1]\). We further define
\[
\tau_{ij}(k)
=
\E_{C\sim q_\psi(\cdot\mid\vx_i)}
\left[
    \alpha_{ij}^{\min\{k,C\}}
\right],
\qquad
\tau_{ij}(0)=1.
\]

For Proposition~\ref{prop:greedy}, the same argument applies when the
exact summary \(\alpha_{ij}\) is replaced by its network approximation
\([L_\eta(\vx_i)]_j\), because the sigmoid output ensures that
\([L_\eta(\vx_i)]_j\in(0,1)\).

\subsection{Proof of Proposition~\ref{prop:closed-form}}

Fix a current frontier
\(\frontier=\{\vx_1,\ldots,\vx_n\}\), remaining budget \(r\), round
budget \(s\le r\), and allocation \(\vk\) satisfying
\(\sum_{i=1}^n k_i=s\). For referrer \(i\), let
\[
N_i=\min\{k_i,C_i\}
\]
denote its number of recruits, and let
\(\{\vY_{i\ell}\}_{\ell=1}^{N_i}\) denote their covariates.

Conditional on the current frontier, the capacities \(C_i\) are
independent across referrers. Conditional on \(C_i\), the recruit
covariates generated by referrer \(i\) are i.i.d.\ from
\(G_\theta(\cdot\mid\vx_i)\), and recruit-generation processes are
independent across referrers.

For each feature \(j\), define
\[
S'_j
\coloneqq
\sum_{i=1}^n
\sum_{\ell=1}^{N_i}
h_{\phi,j}(\vY_{i\ell}).
\]
By Equation~\eqref{eq:structured-value} and linearity of expectation,
\begin{align*}
\E\!\left[
    V_\phi(r-s,\frontier')
\right]
&=
\E\!\left[
    \sum_{j=1}^d
    w_{\phi,j}(r-s)
    \left(
        1-e^{-S'_j}
    \right)
\right]
\\
&=
\sum_{j=1}^d
w_{\phi,j}(r-s)
\left(
    1-\E\!\left[e^{-S'_j}\right]
\right).
\end{align*}

Conditional independence across referrers gives
\begin{equation}
\label{eq:product-factor}
\E\!\left[e^{-S'_j}\right]
=
\prod_{i=1}^n
\E\!\left[
    \exp\left(
        -\sum_{\ell=1}^{N_i}
        h_{\phi,j}(\vY_{i\ell})
    \right)
\right].
\end{equation}

For each referrer \(i\), conditioning on \(N_i\) yields
\begin{align*}
&
\E\!\left[
    \exp\left(
        -\sum_{\ell=1}^{N_i}
        h_{\phi,j}(\vY_{i\ell})
    \right)
    \,\middle|\,
    N_i
\right]
\\
&\qquad=
\prod_{\ell=1}^{N_i}
\E_{\vy\sim G_\theta(\cdot\mid\vx_i)}
\left[
    \exp\bigl(-h_{\phi,j}(\vy)\bigr)
\right]
\\
&\qquad=
\alpha_{ij}^{N_i}.
\end{align*}
The empty product is interpreted as one when \(N_i=0\).

Averaging over \(N_i=\min\{k_i,C_i\}\) gives
\begin{align*}
&
\E\!\left[
    \exp\left(
        -\sum_{\ell=1}^{N_i}
        h_{\phi,j}(\vY_{i\ell})
    \right)
\right]
\\
&\qquad=
\E_{C_i\sim q_\psi(\cdot\mid\vx_i)}
\left[
    \alpha_{ij}^{\min\{k_i,C_i\}}
\right]
\\
&\qquad=
\tau_{ij}(k_i).
\end{align*}
Substituting into Equation~\eqref{eq:product-factor} gives
\[
\E\!\left[e^{-S'_j}\right]
=
\prod_{i=1}^n\tau_{ij}(k_i).
\]
Therefore,
\[
\E\!\left[
    V_\phi(r-s,\frontier')
\right]
=
\sum_{j=1}^d
w_{\phi,j}(r-s)
\left(
    1-
    \prod_{i=1}^n\tau_{ij}(k_i)
\right),
\]
which is the product-form expression in
Proposition~\ref{prop:closed-form}. \qed

\subsection{Proof of Proposition~\ref{prop:greedy}}

Fix a candidate round budget \(s\). For the analysis, extend
\(f_s\) to all allocations \(\vk\in\N^n\) using
\[
f_s(\vk)
=
A(\vk)
+
\gamma
\sum_{j=1}^d
w_{\phi,j}(r-s)B_j(\vk),
\]
where
\[
A(\vk)
=
\sum_{i=1}^n
\sum_{\ell=1}^{k_i}
p_{\psi,i}(\ell)
\]
and
\[
B_j(\vk)
=
1-\prod_{i=1}^n\tau_{ij}(k_i).
\]
The continuation budget \(r-s\) is held fixed throughout this
extension. For allocations satisfying \(\|\vk\|_1=s\), this expression
coincides with the deterministic approximate Q-value
\(\widetilde Q_\phi(r,\frontier,s,\vk)\).

We show that \(f_s\) is monotone and satisfies diminishing returns on
\(\N^n\): for any \(\vk\le\vk'\) coordinate-wise and any coordinate
\(i\),
\[
f_s(\vk+\ve_i)-f_s(\vk)
\ge
f_s(\vk'+\ve_i)-f_s(\vk').
\]

\paragraph{Monotonicity.}

Because \(\alpha_{ij}\in(0,1]\), the quantity
\(\alpha_{ij}^{\min\{k,C\}}\) is nonincreasing in \(k\) for every fixed
\(C\). Therefore,
\[
0<\tau_{ij}(k)\le1
\]
and \(\tau_{ij}(k)\) is nonincreasing in \(k\). It follows that
\(B_j(\vk)\) is nondecreasing in every coordinate.

Moreover,
\[
A(\vk+\ve_i)-A(\vk)
=
p_{\psi,i}(k_i+1)
\ge0.
\]
Since \(w_{\phi,j}(r-s)\ge0\) and \(\gamma\ge0\),
\[
f_s(\vk+\ve_i)-f_s(\vk)
\ge0.
\]
Thus, \(f_s\) is monotone.

\paragraph{Diminishing returns of the immediate-reward term.}

Fix \(\vk\le\vk'\) coordinate-wise and a coordinate \(i\). The marginal
gain of the immediate-reward term is
\[
A(\vk+\ve_i)-A(\vk)
=
p_{\psi,i}(k_i+1).
\]
Because a survival probability is nonincreasing in its threshold and
\(k_i\le k_i'\),
\[
p_{\psi,i}(k_i+1)
\ge
p_{\psi,i}(k_i'+1).
\]
Therefore,
\[
A(\vk+\ve_i)-A(\vk)
\ge
A(\vk'+\ve_i)-A(\vk'),
\]
so \(A\) satisfies diminishing returns.

\paragraph{Diminishing returns of the future-value term.}

For each feature \(j\), define
\[
u_j(\vk)
=
\prod_{i'=1}^n
\tau_{i'j}(k_{i'})
\]
and
\[
\rho_{ij}(k)
=
1-
\frac{\tau_{ij}(k+1)}
     {\tau_{ij}(k)}.
\]
Because \(\tau_{ij}(k)\) is positive and nonincreasing,
\[
0\le \rho_{ij}(k)<1.
\]
The marginal gain of \(B_j\) from incrementing coordinate \(i\) is
\begin{align}
B_j(\vk+\ve_i)-B_j(\vk)
&=
\prod_{i'=1}^n\tau_{i'j}(k_{i'})
-
\tau_{ij}(k_i+1)
\prod_{i'\ne i}\tau_{i'j}(k_{i'})
\nonumber
\\
&=
u_j(\vk)
\left(
    1-
    \frac{\tau_{ij}(k_i+1)}
         {\tau_{ij}(k_i)}
\right)
\nonumber
\\
&=
u_j(\vk)\rho_{ij}(k_i).
\label{eq:Bj-marginal}
\end{align}

Since each \(\tau_{i'j}\) is nonincreasing,
\[
\vk\le\vk'
\quad\Longrightarrow\quad
u_j(\vk)\ge u_j(\vk').
\]

It remains to show that \(\rho_{ij}(k)\) is nonincreasing in \(k\).
For brevity, write
\[
\alpha=\alpha_{ij},
\qquad
\tau(k)=\tau_{ij}(k).
\]
By definition,
\[
\tau(k)
=
\E\!\left[
    \alpha^{\min\{k,C\}}
\right].
\]
Splitting the expectation according to \(C\le k\) and \(C\ge k+1\)
gives
\begin{align}
\tau(k+1)-\alpha\tau(k)
&=
\E\!\left[
    \alpha^{\min\{k+1,C\}}
    -
    \alpha^{1+\min\{k,C\}}
\right]
\nonumber
\\
&=
(1-\alpha)
\E\!\left[
    \alpha^C\mathbf 1\{C\le k\}
\right]
\nonumber
\\
&\eqqcolon d(k).
\label{eq:ratio-id}
\end{align}
The quantity \(d(k)\) is nonnegative and nondecreasing in \(k\),
whereas \(\tau(k)\) is positive and nonincreasing.

For any \(k'\ge k\),
\[
\frac{d(k')}{\tau(k')}
\ge
\frac{d(k)}{\tau(k')}
\ge
\frac{d(k)}{\tau(k)}.
\]
The first inequality uses \(d(k')\ge d(k)\), and the second uses
\(d(k)\ge0\) and \(\tau(k')\le\tau(k)\). Therefore,
\(d(k)/\tau(k)\) is nondecreasing in \(k\).

Dividing Equation~\eqref{eq:ratio-id} by \(\tau(k)>0\) gives
\[
\frac{\tau(k+1)}{\tau(k)}
=
\alpha+\frac{d(k)}{\tau(k)}.
\]
Hence, the ratio
\[
\frac{\tau(k+1)}{\tau(k)}
\]
is nondecreasing in \(k\), and consequently
\[
\rho_{ij}(k)
=
1-
\frac{\tau_{ij}(k+1)}
     {\tau_{ij}(k)}
\]
is nonincreasing in \(k\). Thus, whenever \(k_i\le k_i'\),
\[
\rho_{ij}(k_i)
\ge
\rho_{ij}(k_i').
\]

Combining this inequality with
\(u_j(\vk)\ge u_j(\vk')\), and using the nonnegativity of both factors,
Equation~\eqref{eq:Bj-marginal} gives
\begin{align*}
B_j(\vk+\ve_i)-B_j(\vk)
&=
u_j(\vk)\rho_{ij}(k_i)
\\
&\ge
u_j(\vk')\rho_{ij}(k_i')
\\
&=
B_j(\vk'+\ve_i)-B_j(\vk').
\end{align*}
Therefore, every \(B_j\) satisfies diminishing returns. Since
\(\gamma w_{\phi,j}(r-s)\ge0\), the nonnegative weighted sum of the
\(B_j\)'s also satisfies diminishing returns. Together with the result
for \(A\), this proves that \(f_s\) satisfies diminishing returns.

\paragraph{Greedy approximation guarantee.}

The case \(s=0\) is immediate. Suppose \(s\ge1\). Let
\(\vk^{(t)}\) denote the greedy allocation after \(t\) vouchers, so
that
\[
\|\vk^{(t)}\|_1=t,
\]
and let \(\vk^\star\) be an optimal fixed-budget allocation satisfying
\[
\|\vk^\star\|_1=s.
\]

By monotonicity, any partial allocation \(\vk^{(t)}\) can be completed
to an allocation of size \(s\) without decreasing its value. Since
\(\vk^\star\) is optimal among allocations of size \(s\), it follows
that
\[
f_s(\vk^{(t)})\le f_s(\vk^\star).
\]

Furthermore, monotonicity gives
\[
f_s(\vk^\star)
\le
f_s(\vk^{(t)}\vee\vk^\star),
\]
where \(\vee\) denotes the coordinate-wise maximum. Hence,
\begin{align*}
f_s(\vk^\star)-f_s(\vk^{(t)})
\le{}&
f_s(\vk^{(t)}\vee\vk^\star)
-
f_s(\vk^{(t)}).
\end{align*}

Starting from \(\vk^{(t)}\), the vector
\(\vk^{(t)}\vee\vk^\star\) can be reached by adding
\[
\bigl(k_i^\star-k_i^{(t)}\bigr)_+
\]
units to each coordinate \(i\). By repeatedly applying diminishing
returns, every such unit has marginal gain no larger than the first
marginal at \(\vk^{(t)}\). Therefore,
\begin{align*}
&
f_s(\vk^{(t)}\vee\vk^\star)
-
f_s(\vk^{(t)})
\\
&\quad\le
\sum_{i=1}^n
\bigl(k_i^\star-k_i^{(t)}\bigr)_+
\left[
    f_s(\vk^{(t)}+\ve_i)
    -
    f_s(\vk^{(t)})
\right]
\\
&\quad\le
\sum_{i=1}^n
k_i^\star
\left[
    f_s(\vk^{(t)}+\ve_i)
    -
    f_s(\vk^{(t)})
\right].
\end{align*}

At iteration \(t+1\), greedy selects a coordinate with the largest
available marginal gain. Thus, for every \(i\),
\[
f_s(\vk^{(t)}+\ve_i)-f_s(\vk^{(t)})
\le
f_s(\vk^{(t+1)})-f_s(\vk^{(t)}).
\]
Since \(\sum_i k_i^\star=s\), we obtain
\begin{align*}
f_s(\vk^\star)-f_s(\vk^{(t)})
\le
s
\left[
    f_s(\vk^{(t+1)})
    -
    f_s(\vk^{(t)})
\right].
\end{align*}
Equivalently,
\[
f_s(\vk^{(t+1)})-f_s(\vk^{(t)})
\ge
\frac{1}{s}
\left[
    f_s(\vk^\star)-f_s(\vk^{(t)})
\right].
\]
Subtracting this gain from the current optimality gap gives
\begin{align*}
f_s(\vk^\star)-f_s(\vk^{(t+1)})
&=
f_s(\vk^\star)-f_s(\vk^{(t)})
\\
&\quad-
\left[
    f_s(\vk^{(t+1)})
    -
    f_s(\vk^{(t)})
\right]
\\
&\le
\left(
    1-\frac{1}{s}
\right)
\left[
    f_s(\vk^\star)-f_s(\vk^{(t)})
\right].
\end{align*}
Iterating for \(t=0,\ldots,s-1\), and using
\[
f_s(\mathbf 0)=0,
\]
gives
\begin{align*}
f_s(\vk^\star)-f_s(\vk^{(s)})
&\le
\left(
    1-\frac{1}{s}
\right)^s
f_s(\vk^\star).
\end{align*}
Therefore,
\begin{align*}
f_s(\vk^{(s)})
&\ge
\left[
    1-
    \left(
        1-\frac{1}{s}
    \right)^s
\right]
f_s(\vk^\star)
\\
&\ge
(1-1/e)f_s(\vk^\star).
\end{align*}
Since \(\vk^{(s)}\) is the allocation returned by marginal greedy, this
proves the claimed fixed-budget approximation guarantee. \qed

\begin{table*}[t]
\centering
\caption{
Cumulative rewards (mean $\pm$ standard error), with the highest mean
in each column shown in bold. The shifted setting uses a separately
trained covariate model and $\varepsilon_c = 0.2$. We use $B = 200$,
$|F_0| = 10$, $\gamma = 0.9$, and evaluate over 50 episodes for ICPSR
and 30 episodes for FHI~360.
}
\label{tab:additional-gamma-09}

\resizebox{\textwidth}{!}{%
\begin{tabular}{@{}lcccc@{\hspace{10pt}}cccc@{}}
\toprule
& \multicolumn{4}{c}{ICPSR}
& \multicolumn{4}{c}{FHI~360} \\
\cmidrule(lr){2-5}
\cmidrule(lr){6-9}

& \multicolumn{2}{c}{Discounted}
& \multicolumn{2}{c}{Undiscounted}
& \multicolumn{2}{c}{Discounted}
& \multicolumn{2}{c}{Undiscounted} \\
\cmidrule(lr){2-3}
\cmidrule(lr){4-5}
\cmidrule(lr){6-7}
\cmidrule(lr){8-9}

Method
& Clean & Shift
& Clean & Shift
& Clean & Shift
& Clean & Shift \\
\midrule

Random
& 57.4 \std{ 3.4}
& 64.0 \std{ 3.6}
& 65.9 \std{ 4.0}
& 73.2 \std{ 4.3}
& 28.3 \std{ 2.1}
& 47.4 \std{ 2.8}
& 31.6 \std{ 2.5}
& 53.7 \std{ 3.3} \\

Constant-2
& 115.2 \std{ 1.3}
& 100.3 \std{ 1.8}
& 151.4 \std{ 1.4}
& 135.5 \std{ 1.5}
& 51.5 \std{ 2.0}
& 55.0 \std{ 2.8}
& 84.7 \std{ 3.1}
& 82.2 \std{ 4.6} \\

Constant-3
& 112.0 \std{ 2.1}
& 101.4 \std{ 1.5}
& 134.3 \std{ 2.3}
& 120.1 \std{ 1.7}
& 49.5 \std{ 1.6}
& 57.5 \std{ 1.5}
& 65.2 \std{ 1.7}
& 72.9 \std{ 1.1} \\

Constant-5
& 92.9 \std{ 2.0}
& 87.7 \std{ 2.3}
& 104.5 \std{ 2.2}
& 97.7 \std{ 2.5}
& 36.8 \std{ 1.1}
& 46.8 \std{ 1.5}
& 41.8 \std{ 1.1}
& 52.4 \std{ 1.6} \\

Constant-10
& 53.1 \std{ 2.0}
& 58.0 \std{ 2.2}
& 56.7 \std{ 2.2}
& 61.6 \std{ 2.4}
& 22.0 \std{ 0.9}
& 36.0 \std{ 1.6}
& 22.9 \std{ 0.9}
& 37.6 \std{ 1.7} \\

Budget DQN
& 132.4 \std{ 1.7}
& 121.0 \std{ 2.0}
& 151.6 \std{ 1.7}
& 139.0 \std{ 2.2}
& 71.6 \std{ 4.1}
& \textbf{73.7} \std{ 4.1}
& 102.0 \std{ 4.8}
& 105.5 \std{ 4.5} \\

Factorized RL
& 115.8 \std{ 3.2}
& 106.3 \std{ 2.9}
& 156.2 \std{ 1.7}
& 139.0 \std{ 2.2}
& 34.4 \std{ 1.4}
& 46.1 \std{ 1.7}
& 49.6 \std{ 2.3}
& 65.5 \std{ 2.6} \\

IID Pop DP
& 157.3 \std{ 1.2}
& 141.5 \std{ 2.1}
& 183.3 \std{ 0.8}
& 166.4 \std{ 1.5}
& 68.5 \std{ 3.9}
& 71.7 \std{ 3.1}
& 92.3 \std{ 4.3}
& 95.4 \std{ 3.2} \\

\midrule
\textbf{GFP (ours)}
& \textbf{160.1} \std{ 1.9}
& \textbf{144.4} \std{ 1.8}
& \textbf{192.7} \std{ 1.2}
& \textbf{174.0} \std{ 1.5}
& \textbf{77.8} \std{ 3.6}
& 73.6 \std{ 3.4}
& \textbf{114.3} \std{ 4.3}
& \textbf{107.9} \std{ 4.4} \\
\bottomrule
\end{tabular}%
}
\end{table*}

\begin{table*}[t]
\centering
\caption{
Total recruitment at \(\gamma=1.0\)
(mean \(\pm\) standard error).
Because \(\gamma=1.0\), discounted cumulative reward is identical to
total recruitment. The shifted setting uses a separately trained
covariate model and count-contamination level
\(\varepsilon_c=0.2\).
We use \(B=200\), \(|\frontier_0|=10\), and evaluate over \(50\)
episodes for ICPSR and \(30\) episodes for FHI~360.
The highest mean in each column is shown in bold.
}
\label{tab:additional-gamma-10}

\begin{tabular}{@{}lcccc@{}}
\toprule
& \multicolumn{2}{c}{ICPSR}
& \multicolumn{2}{c}{FHI~360} \\
\cmidrule(lr){2-3}
\cmidrule(lr){4-5}

Method
& Clean & Shifted
& Clean & Shifted \\
\midrule

Random
& 65.9 \std{ 4.0}
& 73.2 \std{ 4.3}
& 31.6 \std{ 2.5}
& 53.7 \std{ 3.3} \\

Constant-2
& 151.4 \std{ 1.4}
& 135.5 \std{ 1.5}
& 84.7 \std{ 3.1}
& 82.2 \std{ 4.6} \\

Constant-3
& 134.3 \std{ 2.3}
& 120.1 \std{ 1.7}
& 65.2 \std{ 1.7}
& 72.9 \std{ 1.1} \\

Constant-5
& 104.5 \std{ 2.2}
& 97.7 \std{ 2.5}
& 41.8 \std{ 1.1}
& 52.4 \std{ 1.6} \\

Constant-10
& 56.7 \std{ 2.2}
& 61.6 \std{ 2.4}
& 22.9 \std{ 0.9}
& 37.6 \std{ 1.7} \\

Budget-DQN
& 151.8 \std{ 2.0}
& 137.7 \std{ 2.2}
& 101.0 \std{ 4.5}
& 100.3 \std{ 4.5} \\

Factorized RL
& 169.1 \std{ 6.5}
& 154.7 \std{ 5.1}
& 30.8 \std{ 2.4}
& 61.2 \std{ 3.8} \\

IID-Population DP
& 193.6 \std{ 0.5}
& 170.3 \std{ 3.6}
& 98.4 \std{ 4.6}
& 102.7 \std{ 3.8} \\

\midrule
\textbf{GFP (ours)}
& \textbf{199.5} \std{ 0.2}
& \textbf{173.2} \std{ 4.6}
& \textbf{126.0} \std{ 5.4}
& \textbf{110.9} \std{ 4.6} \\

\bottomrule
\end{tabular}
\end{table*}

\section{Additional  Results}
\label{sect:appendix:additional-results}

\paragraph{Main results.} To complement the main results at \(\gamma=0.99\),
Tables~\ref{tab:additional-gamma-09} and
\ref{tab:additional-gamma-10} report results for
\(\gamma=0.9\) and \(\gamma=1.0\), respectively.
GFP achieves the highest mean in every setting except for discounted
reward under shifted FHI~360 dynamics at \(\gamma=0.9\), where it nearly
matches Budget-DQN (\(73.6\) versus \(73.7\)) while achieving higher
total recruitment (\(107.9\) versus \(105.5\)).
At \(\gamma=1.0\), discounted cumulative reward and total recruitment
coincide by definition. Overall, GFP performs consistently across
discount factors, datasets, and test conditions.

\paragraph{Robustness to count contamination.}
We further evaluate robustness to test-time shifts in recruitment counts
on FHI~360. As shown in Table~\ref{tab:noise-sensitivity-fhi360}, GFP achieves
the highest mean total recruitment at every contamination level
\(\varepsilon_c\in\{0.1,0.2,0.3,0.4\}\). Relative to the strongest
baseline in each column, GFP improves mean total recruitment by \(8.9\),
\(12.2\), \(10.1\), and \(2.6\), respectively. The strongest baseline is
Budget-DQN at \(\varepsilon_c\in\{0.1,0.2\}\) and IID-Population DP at
\(\varepsilon_c\in\{0.3,0.4\}\).

As the contamination level increases from \(0.1\) to \(0.4\), GFP's mean
total recruitment decreases from \(114.7\) to \(106.9\), while remaining
above both Budget-DQN and IID-Population DP throughout. The smaller
margin at \(\varepsilon_c=0.4\) indicates that the methods become more
competitive under stronger count shifts, but GFP still retains the
highest mean performance.

\begin{table}[htbp]
\centering
\small
\caption{
Total recruitment under test-time count contamination on FHI~360 at
\(\gamma=0.99\). The shifted covariate model is held fixed across
contamination levels, while only \(\varepsilon_c\) varies. Entries are
mean \(\pm\) standard error, and the highest mean in each column is shown
in bold.
}
\label{tab:noise-sensitivity-fhi360}
\resizebox{\columnwidth}{!}{%
\begin{tabular}{@{}lcccc@{}}
\toprule
Method
& \(\varepsilon_c{=}0.1\)
& \(\varepsilon_c{=}0.2\)
& \(\varepsilon_c{=}0.3\)
& \(\varepsilon_c{=}0.4\) \\
\midrule

Random
& 38.9 \std{ 2.7}
& 53.7 \std{ 3.3}
& 56.7 \std{ 3.0}
& 64.6 \std{ 3.4} \\

Constant-2
& 81.6 \std{ 3.6}
& 82.2 \std{ 4.6}
& 87.5 \std{ 3.6}
& 80.3 \std{ 4.2} \\

Constant-3
& 66.4 \std{ 1.3}
& 72.9 \std{ 1.1}
& 75.6 \std{ 1.2}
& 76.5 \std{ 2.5} \\

Constant-5
& 47.0 \std{ 1.1}
& 52.4 \std{ 1.6}
& 60.1 \std{ 1.9}
& 66.7 \std{ 2.1} \\

Constant-10
& 30.3 \std{ 1.4}
& 37.6 \std{ 1.7}
& 45.4 \std{ 2.3}
& 56.1 \std{ 3.1} \\

Budget-DQN
& 105.8 \std{ 2.6}
& 100.6 \std{ 4.8}
& 99.1 \std{ 3.2}
& 100.5 \std{ 3.7} \\

Factorized RL
& 46.9 \std{ 3.9}
& 48.5 \std{ 3.7}
& 62.2 \std{ 3.2}
& 68.0 \std{ 3.5} \\

IID-Population DP
& 102.5 \std{ 3.1}
& 97.7 \std{ 3.2}
& 102.7 \std{ 2.7}
& 104.3 \std{ 2.5} \\

\midrule
\textbf{GFP (ours)}
& \textbf{114.7} \std{ 5.4}
& \textbf{112.8} \std{ 3.7}
& \textbf{112.8} \std{ 3.3}
& \textbf{106.9} \std{ 4.5} \\

\bottomrule
\end{tabular}%
}
\end{table}

\section{Experimental Details for the Data-Calibrated Environments}
\label{app:environment-details}

\paragraph{ICPSR.}
In ICPSR, directed edges represent observed social or contact ties
rather than records of issued referral vouchers. We therefore use node
out-degree only as a network-based proxy for latent referral capacity
when constructing the data-calibrated simulator.

\paragraph{FHI~360.}
In the FHI~360 dataset, each observed child corresponds to a successful
coupon-mediated referral. However, the number of vouchers issued to a
participant, \(k_i\), is unavailable. Consequently, the observed child
count does not reveal whether the participant's referral capacity was
fully observed or censored by the available vouchers. To construct
training data for the censored count model, we use node out-degree
\(\widetilde C_i\) as a proxy for latent capacity, sample a
counterfactual allocation \(k_i\), and define
\[
y_i=\min\{k_i,\widetilde C_i\}.
\]
This procedure produces synthetic censored observations with the same
observation structure assumed by the count model.

\subsection{Data-Calibrated Environment Specifications}
\label{app:real-env-details}

\paragraph{Covariate schemas.}
The FHI~360 environment uses the \(101\)-dimensional one-hot covariate
representation listed in Table~\ref{tab:covariate_schema_epoa}. It also
contains \(17\) categorical fields, with exactly one active entry in
each field.

The ICPSR environment uses the \(72\)-dimensional one-hot covariate
representation listed in Table~\ref{tab:covariate_schema}. The same
schema is also used in the additional synthetic environment. It
contains \(17\) categorical fields, each represented by a contiguous
one-hot block with exactly one active entry.

\begin{table}[h]
\centering
\small
\caption{
The \(17\) categorical fields composing the
\(101\)-dimensional FHI~360 covariate representation.
}
\label{tab:covariate_schema_epoa}
\begin{tabular}{llr}
\toprule
Field & Description & Size \\
\midrule
\textsc{Gender}  &  Gender of participant    & 4 \\
\textsc{Population} & Detailed Population group & 24 \\
\textsc{Sex}    & Sex at birth           & 3 \\
\textsc{KPType} & Key-population type & 10 \\
\textsc{Age} & Age at enrollment & 9 \\
\textsc{LastTest} & Approximate time since last HIV test & 5 \\
\textsc{RecentSex} & Sexual activity in the last month & 3 \\
\textsc{RecentTest} & HIV testing in the last six months & 3 \\
\textsc{SexExch} & Exchange of sex for money or goods & 3 \\
\textsc{Hierarchy} & Level of hierarchy & 3 \\
\textsc{Region} & Geographic region & 13 \\
\textsc{CondomUse} & Condom use during sex & 2 \\
\textsc{CondomFreq} & Frequency of condom use & 5 \\
\textsc{SubstFreq} & Drug or alcohol use during sex & 5 \\
\textsc{TBSusp} & Suspected tuberculosis & 3 \\
\textsc{STISymp} & Reported STI symptoms & 3 \\
\textsc{GBV} & Reported gender-based violence & 3 \\
\midrule
Total & & 101 \\
\bottomrule
\end{tabular}
\end{table}

\begin{table}[h]
\centering
\small
\caption{
The \(17\) categorical fields composing the
\(72\)-dimensional ICPSR covariate representation.
}
\label{tab:covariate_schema}
\begin{tabular}{llr}
\toprule
Field & Description & Size \\
\midrule
\textsc{Local}   & Locality            & 4 \\
\textsc{Race}    & Race                & 7 \\
\textsc{Ethn}    & Ethnicity           & 4 \\
\textsc{Sex}     & Sex                 & 3 \\
\textsc{Orient}  & Sexual orientation  & 6 \\
\textsc{Behav}   & Sexual behavior     & 3 \\
\textsc{Pro}     & Sex-work profession & 4 \\
\textsc{Pimp}    & Pimp status         & 4 \\
\textsc{John}    & John status         & 4 \\
\textsc{Dealer}  & Drug dealing        & 4 \\
\textsc{Drugman} & Drug managing       & 4 \\
\textsc{Thief}   & Theft               & 4 \\
\textsc{Retired} & Retired             & 4 \\
\textsc{Hwife}   & Homemaker           & 4 \\
\textsc{Disable} & Disability          & 5 \\
\textsc{Unemp}   & Unemployed          & 4 \\
\textsc{Streets} & Street-involved     & 4 \\
\midrule
Total & & 72 \\
\bottomrule
\end{tabular}
\end{table}

\subsection{Training Details}
\label{app:real-training}

We use the same model classes, network architectures, and optimization
settings across the ICPSR and FHI~360 experiments. All covariate pools,
count-model samples, recruit-covariate pairs, and planning transitions
are constructed separately from the corresponding dataset-specific
planning environment. Shared hyperparameters are reported in
Table~\ref{tab:hyperparams_FHISTI}.

\paragraph{Learned referral dynamics.}
We fit the count model \(q_\psi\) on censored triples
\((\vx_i,k_i,y_i)\). Its Poisson rate is parameterized by a
two-hidden-layer MLP with ReLU activations and a softplus output, and
the model is optimized using the censored likelihood defined in
Section~\ref{sec:method}.

We fit the recruit-covariate model \(G_\theta\) on observed
referrer--recruit covariate pairs from the corresponding dataset. We
instantiate \(G_\theta\) as a conditional DDPM
\cite{ho2020denoising}. Its noise-prediction network is a
three-hidden-layer MLP with GELU activations, conditioned on the
referrer covariates, the noised recruit covariates, and a
\(16\)-dimensional sinusoidal timestep embedding. To obtain a valid
categorical covariate vector from each continuous DDPM sample, we take
the argmax separately within each categorical field.

\paragraph{Reinforcement learning baselines.}
Budget-DQN represents the variable-size frontier using a DeepSets
encoder. The resulting frontier representation and the remaining
budget are passed to a two-hidden-layer Q-network, which outputs one
Q-value for each feasible round budget. After selecting a round budget,
Budget-DQN ranks current referrers by their predicted recruitment counts
and allocates vouchers greedily according to this ranking.

Factorized RL uses the same DeepSets frontier encoder, followed by
three two-hidden-layer prediction heads for the round budget, the
number of active referrers, and node-level allocation scores. It first
selects the highest-valued feasible round budget and support size, then
chooses the corresponding highest-scoring referrers. The selected
budget is distributed according to their softmax-normalized scores and
converted into nonnegative integer allocations while preserving the
total round budget.

Both baselines are trained in the corresponding dataset-specific
planning environment using experience replay, a target network, and
\(\varepsilon\)-greedy exploration. Their shared hyperparameters are
reported in Table~\ref{tab:hyperparams_FHISTI}.

\paragraph{GFP training.}
The structured value surrogate uses latent dimension \(d=32\) and
hidden dimension \(64\). The feature network \(h_\phi\) is a
two-hidden-layer MLP, and the budget-weight network is a
one-hidden-layer MLP.

The referrer-summary network \(L_\eta\) approximates
\(\boldsymbol{\alpha}_{\theta,\phi}(\vx)\) and is implemented as a
two-hidden-layer MLP with hidden dimension \(64\) and output dimension
\(d\). We initialize \(L_\eta\) using \(256\) referrer covariates and
\(64\) recruit samples per referrer, optimizing it for \(200\) gradient
steps with Adam, learning rate \(10^{-3}\), and batch size \(128\).
During fitted value iteration, we refresh \(L_\eta\) every \(25\)
iterations using the current feature representation \(h_\phi\) and new
recruit samples from the fixed generative model \(G_\theta\).

We train the value surrogate by fitted value iteration for \(200\)
iterations with batch size \(16\) and learning rate \(10^{-3}\). The
training pool contains \(256\) frontier--budget states collected from
\(64\) random-rollout episodes in the corresponding planning
environment.

\paragraph{IID-Population DP~\cite{pan2026adaptive}.}
We draw \(M=1{,}024\) covariate vectors
\(\{\vx^{(m)}\}_{m=1}^M\) from the planning pool and construct the
marginal population-level count distribution
\[
\overline q_\psi(c)
=
\frac{1}{M}
\sum_{m=1}^M q_\psi\!\left(c\mid\vx^{(m)}\right).
\]
The baseline then performs standard backward induction over
remaining-budget and frontier-size pairs up to the full budget \(B\).

\begin{table*}[h]
\centering
\small
\caption{
Shared hyperparameters for the ICPSR and FHI~360 data-calibrated
environments. All optimizers are Adam.
}
\label{tab:hyperparams_FHISTI}
\begin{tabular}{llr}
\toprule
Component & Parameter & Value \\
\midrule

\multicolumn{3}{l}{\textit{Learned count model \(q_\psi\)}} \\
& Training samples & \(4{,}096\) \\
& Maximum allocation & 10 \\
& Epochs / batch / lr & 200 / 128 / \(10^{-3}\) \\

\midrule
\multicolumn{3}{l}{\textit{Learned covariate model \(G_\theta\) (DDPM)}} \\
& Training pairs & \(80\%\) of available pairs \\
& Epochs / batch / lr & \(4{,}000\) / 32 / \(10^{-3}\) \\
& Hidden dimension / diffusion steps & 512 / 100 \\

\midrule
\multicolumn{3}{l}{\textit{Budget-DQN and Factorized RL}} \\
& Training episodes & 200 \\
& Replay-buffer capacity & \(100{,}000\) \\
& Batch / hidden / lr & 64 / 128 / \(10^{-3}\) \\
& Training discount factor & Matched to evaluation \\
& \(\varepsilon\) schedule & \(0.20 \rightarrow 0.05\) \\

\midrule
\multicolumn{3}{l}{\textit{GFP value surrogate}} \\
& Latent dimension \(d\) & 32 \\
& Hidden dimension & 64 \\
& Iterations / batch / lr & 200 / 16 / \(10^{-3}\) \\
& State-pool size & 256 \\
& Random-rollout episodes & 64 \\

\midrule
\multicolumn{3}{l}{\textit{GFP referrer-summary network \(L_\eta\)}} \\
& Training referrers & 256 \\
& Recruit samples / referrer & 64 \\
& Refresh interval & 25 iterations \\
& Steps / batch / lr & 200 / 128 / \(10^{-3}\) \\

\midrule
\multicolumn{3}{l}{\textit{IID-Population DP}} \\
& Population sample size & \(1{,}024\) \\

\bottomrule
\end{tabular}
\end{table*}

\section{Additional Synthetic Environment}
\label{subsec:env}

We additionally evaluate all methods in a synthetic environment designed
to isolate the benefit of covariate-aware planning under controlled
ground-truth dynamics. The environment follows the sequential structure
of respondent-driven sampling, while its covariate schema and
referrer--recruit dependence are calibrated to the ICPSR~22140
dataset. Unlike the data-calibrated environments in the main
experiments, the referral-capacity dynamics are specified
parametrically and are known only to the simulator.

Each individual is represented by a \(p=72\)-dimensional one-hot
covariate vector comprising \(K=17\) categorical demographic,
behavioral, and socioeconomic fields from the ICPSR~22140
schema~\cite{morris2011hiv}. The oracle referral-capacity model is a
covariate-dependent Poisson model whose rates are calibrated to have
mean \(\bar{\lambda}=2.5\) over the covariate population. The rates
exhibit a \(5\)--\(10\times\) spread between the highest- and
lowest-capacity individuals, making uniform voucher allocation
substantially suboptimal.

Recruit covariates are generated by a field-wise categorical inheritance
kernel. For each covariate field \(k\), a recruit inherits the
referrer's category with probability \(p_k\); otherwise, its category is
drawn uniformly from the categories in that field. We estimate the
inheritance probabilities from \(73{,}669\) recruiter--recruit dyads in
ICPSR~22140. The resulting values range from
\(p_{\textsc{Sex}}=0.22\) to \(p_{\textsc{Drugman}}=0.98\), capturing
substantial variation in referrer--recruit dependence across fields.

At the beginning of each episode, the initial frontier
\(\frontier_0\) contains \(10\) individuals sampled independently with
replacement from a pre-generated pool of \(N=300\) valid one-hot
covariate vectors. The complete covariate schema, the
inheritance-probability estimator, and the parameters of the oracle
dynamics are provided in Appendix~\ref{app:env-details}.

\subsection{Setup}
\label{subsec:setup}

The synthetic simulator defines the oracle recruitment dynamics
\(q^\star\) and \(G^\star\). These oracle models are used only to
generate evaluation trajectories and are not exposed to the planning
algorithms. Methods that require model fitting or policy training are
developed using learned surrogates \(q_\psi\) and \(G_\theta\), which
together define the \emph{planning environment}. The resulting policies
are then evaluated under the oracle dynamics, which define the
\emph{testing environment}. This protocol prevents methods from
accessing the ground-truth dynamics and evaluates all policies in the
same oracle environment. Full details of training-data generation,
model architectures, and hyperparameters are provided in
Appendix~\ref{app:training}.

\paragraph{Episode protocol.}
Each episode begins with an initial frontier of size
\(|\frontier_0|=10\), a total voucher budget \(B=100\), and a maximum
horizon of \(50\) rounds. An episode terminates early if the budget is
exhausted or the frontier becomes empty. We evaluate four discount
factors,
\[
\gamma \in \{0.9,\,0.95,\,0.99,\,1.0\}.
\]
For each setting, every method uses the same value of \(\gamma\) in its
training or planning objective.

\paragraph{Metrics.}
We report two episode-level metrics, averaged over \(20\) independent
evaluation episodes. Our primary metric is total recruitment,
\[
\sum_t R_t,
\]
which measures the number of individuals reached under the fixed
voucher budget. We additionally report discounted cumulative reward,
\[
\sum_t \gamma^t R_t,
\]
which places greater weight on recruits obtained in earlier rounds.

\subsection{Results}
\label{subsec:results}

\begin{figure*}[t]
  \centering
  \includegraphics[width=\textwidth]{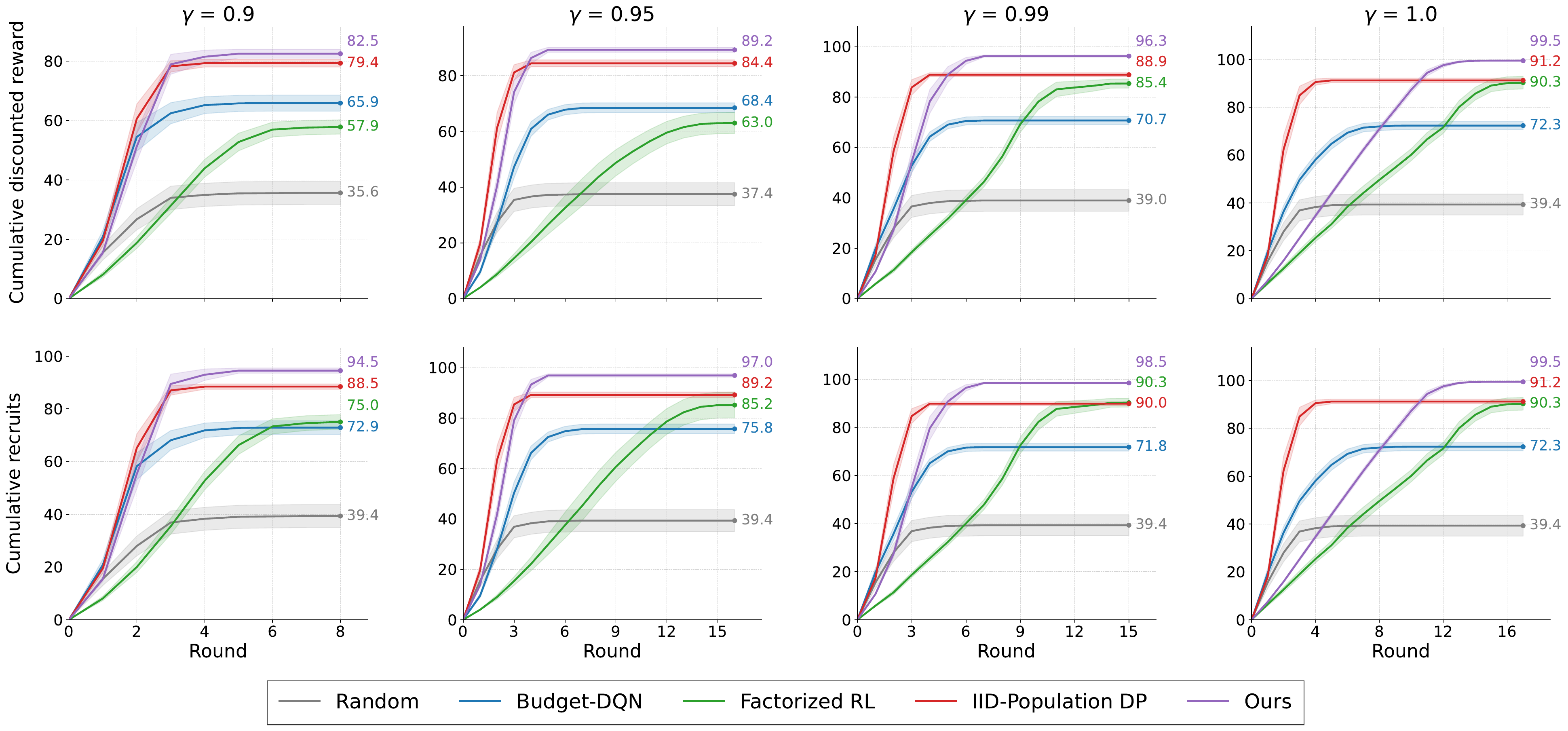}
  \caption{
  Cumulative discounted reward (top) and cumulative recruitment
  (bottom) by round for
  \(\gamma \in \{0.9,\,0.95,\,0.99,\,1.0\}\).
  Lines show means over \(20\) evaluation episodes, and shaded regions
  show \(\pm 1\) standard error.
  }
  \label{fig:overlay}
\end{figure*}

Table~\ref{tab:synth-main} reports total recruitment and discounted
cumulative reward across the four discount factors.
Figure~\ref{fig:overlay} shows the corresponding cumulative
per-round trajectories. Note that we omit the constant strategy for this synthetic environment, as the main text has already shown that it performs relatively poorly compared with the other baselines.

\begin{table}[t]
\centering
\small
\caption{
Results in the additional synthetic environment with \(B=100\) and
\(|\frontier_0|=10\). Entries are mean \(\pm\) standard error over
\(20\) independent evaluation episodes. The highest mean in each
column is shown in bold.
}
\label{tab:synth-main}
\begin{tabular}{lcccc}
\toprule
 & $\gamma{=}0.9$ & $\gamma{=}0.95$ & $\gamma{=}0.99$ & $\gamma{=}1.0$ \\
\midrule
\multicolumn{5}{l}{\emph{Final discounted cumulative reward}} \\
Random                       
& 35.6 \std{ 3.9} 
& 37.4 \std{ 4.1} 
& 39.0 \std{ 4.3} 
& 39.4 \std{ 4.4} \\
Budget-DQN                   
& 65.9 \std{ 2.7} 
& 68.4 \std{ 1.8} 
& 70.7 \std{ 1.6} 
& 72.4 \std{ 1.7} \\
Factorized RL                
& 57.9 \std{ 2.4} 
& 63.0 \std{ 3.8} 
& 85.4 \std{ 1.8} 
& 90.3 \std{ 2.7} \\
IID-Pop. DP            
& 79.4 \std{ 1.3}
& 84.4 \std{ 1.2}
& 88.9 \std{ 0.8} 
& 91.3 \std{ 1.0} \\
GFP (ours)                   
& \textbf{82.5} \std{ 1.5} 
& \textbf{89.2} \std{ 0.9} 
& \textbf{96.3} \std{ 0.4} 
& \textbf{99.5} \std{ 0.2} \\
\midrule
\multicolumn{5}{l}{\emph{Final cumulative recruits}} \\
Random                       
& 39.4 \std{ 4.4} 
& 39.4 \std{ 4.4} 
& 39.4 \std{ 4.4}
& 39.4 \std{ 4.4} \\
Budget-DQN                   
& 72.9 \std{ 2.6} 
& 75.8 \std{ 1.9} 
& 71.9 \std{ 1.6} 
& 72.4 \std{ 1.7} \\
Factorized RL                
& 75.1 \std{ 2.8} 
& 85.2 \std{ 5.1} 
& 90.4 \std{ 1.8} 
& 90.3 \std{ 2.7} \\
IID-Pop. DP            
& 88.5 \std{ 1.0} 
& 89.3 \std{ 1.1} 
& 90.0 \std{ 0.8} 
& 91.3 \std{ 1.0} \\
GFP (ours)                   
& \textbf{94.5} \std{ 1.0}
& \textbf{97.0} \std{ 0.6}
& \textbf{98.6} \std{ 0.3}
& \textbf{99.5} \std{ 0.2} \\
\bottomrule
\end{tabular}
\end{table}

\paragraph{GFP performs best across all discount factors.}
GFP achieves the highest mean total recruitment and discounted
cumulative reward for every value of \(\gamma\), yielding the best
result in all eight metric--discount comparisons. At \(\gamma=1.0\),
GFP recruits \(99.5 \pm 0.2\) individuals under a budget of \(B=100\),
compared with \(91.3 \pm 1.0\) for the strongest baseline in this
setting, IID-Population DP. Thus, GFP converts nearly the entire voucher
budget into successful recruitments. Although the relative ordering of
the baselines varies with \(\gamma\), GFP remains the best-performing
method in every setting.

\paragraph{Covariate composition improves planning.}
IID-Population DP provides the most relevant model-based comparison
because it also performs cross-round budget planning using a surrogate
for the future frontier, but represents that frontier only through its
size. GFP improves total recruitment over IID-Population DP by \(8.2\)
at \(\gamma=1.0\) and \(6.0\) at \(\gamma=0.9\). These improvements are
consistent with the benefit of representing frontier composition when
both referral capacity and recruit covariates depend on the current
referrers.

\paragraph{Exploiting problem structure improves performance.}
The two RL baselines remain below GFP across all discount factors.
Unlike GFP, they do not directly exploit the product-form expected
future value or the diminishing-returns structure of the intra-round
allocation objective. Factorized RL becomes more competitive at larger
values of \(\gamma\), but still does not match GFP in either total
recruitment or discounted reward. Budget-DQN additionally separates its
round-budget decision from its individual-level allocation rule, which
limits its ability to account for the downstream effects of allocating
vouchers to different referrers.

\paragraph{Early recruitment speed does not determine final performance.}
Figure~\ref{fig:overlay} shows that GFP does not always achieve the
largest cumulative recruitment in the earliest rounds. At
\(\gamma=1.0\), IID-Population DP approaches \(90\) recruits within the
first few rounds, while GFP allocates the budget more gradually and
overtakes the baselines later in the episode. Thus, methods that
front-load voucher expenditure can achieve strong early recruitment but
plateau at a lower final value.

GFP's later gains are consistent with its covariate-aware allocation
mechanism. It accounts for both the covariate-conditioned referral
capacity of current participants and the downstream value of the
frontiers generated by their recruits. In contrast, IID-Population DP
cannot distinguish future frontiers of the same size but different
covariate compositions. Random allocation reaches approximately \(39.4\)
total recruits across all settings, illustrating that how vouchers are
allocated can substantially affect how effectively the fixed budget is
converted into successful recruitment.

\section{Experimental Details for the Additional Synthetic Environment}
\label{app:exp-details}

\subsection{Simulation Environment}
\label{app:env-details}

\paragraph{Oracle referral-capacity model.}
For an individual with covariate vector
\(\vx\in\{0,1\}^{72}\), the oracle referral capacity is
\[
C_i\sim\operatorname{Poisson}\!\left(\lambda(\vx_i)\right),
\qquad
\lambda(\vx)
=
\kappa\,\operatorname{softplus}\!\left(\vw^\top\vx\right),
\]
where \(\vw\in\mathbb{R}^{72}\) is sampled once from
\(\mathcal{N}(\mathbf{0},\sigma^2 I)\) and then held fixed across all
episodes. The scaling constant \(\kappa>0\) is chosen so that the mean
rate over the synthetic covariate pool equals the target
\(\bar{\lambda}=2.5\).

\paragraph{Oracle covariate transition model.}
Recruit covariates are generated from referrer covariates using a
field-wise categorical inheritance kernel. Let \(m_k\) denote the
number of categories in field \(k\). Independently for each of the
\(K=17\) fields, the recruit inherits the referrer's category with
probability \(p_k\); otherwise, its category is sampled uniformly from
all \(m_k\) categories:
\[
x'_{\mathrm{group}_k}
\sim
\begin{cases}
\delta_{x_{\mathrm{group}_k}},
    & \text{with probability } p_k, \\[4pt]
\operatorname{Uniform}\!\left(\{1,\ldots,m_k\}\right),
    & \text{with probability } 1-p_k.
\end{cases}
\]
The fields are generated independently conditional on the referrer,
yielding a valid one-hot covariate vector for each recruit.

\paragraph{Empirical inheritance probabilities.}
We estimate \(p_k\) from \(73{,}669\) recruiter--recruit dyads in
ICPSR~22140, using all disease subnetworks and directed edges satisfying
\(\textsc{Ntype1}=1\) and \(\textsc{Ntype2}=3\). Under the transition
kernel above, the probability that the referrer and recruit have the
same category in field \(k\) is
\[
\Pr(\text{match in field }k)
=
p_k+\frac{1-p_k}{m_k}.
\]
Therefore, letting \(\widehat{\mathrm{match}}_k\) denote the empirical
matching frequency, we estimate
\[
p_k
=
\frac{
\widehat{\mathrm{match}}_k-1/m_k
}{
1-1/m_k
}.
\]
The resulting estimates are reported in
Table~\ref{tab:inherit_probs}.

\begin{table}[h]
\centering
\small
\caption{Empirical field-wise inheritance probabilities estimated from
ICPSR~22140 recruiter--recruit dyads.}
\label{tab:inherit_probs}
\begin{tabular}{lr}
\toprule
Field & $p_k$ \\
\midrule
\textsc{Local}   & 0.766 \\
\textsc{Race}    & 0.474 \\
\textsc{Ethn}    & 0.861 \\
\textsc{Sex}     & 0.223 \\
\textsc{Orient}  & 0.744 \\
\textsc{Behav}   & 0.762 \\
\textsc{Pro}     & 0.573 \\
\textsc{Pimp}    & 0.891 \\
\textsc{John}    & 0.680 \\
\textsc{Dealer}  & 0.775 \\
\textsc{Drugman} & 0.979 \\
\textsc{Thief}   & 0.940 \\
\textsc{Retired} & 0.960 \\
\textsc{Hwife}   & 0.861 \\
\textsc{Disable} & 0.865 \\
\textsc{Unemp}   & 0.339 \\
\textsc{Streets} & 0.952 \\
\bottomrule
\end{tabular}
\end{table}

\subsection{Training Details}
\label{app:training}

We use the same model classes and training procedures as in the two
data-calibrated environments, with environment-specific sample sizes
and hyperparameters reported in Table~\ref{tab:hyperparams}. All
training data are generated from the synthetic planning environment;
the oracle testing environment is used only for final evaluation.

\begin{table*}[h]
\centering
\small
\caption{
Hyperparameters for the additional synthetic environment.
All optimizers are Adam.
}
\label{tab:hyperparams}
\begin{tabular}{llr}
\toprule
Component & Parameter & Value \\
\midrule

\multicolumn{3}{l}{\textit{Learned count model \(q_\psi\)}} \\
& Training samples & \(2{,}048\) \\
& Maximum allocation & 10 \\
& Epochs / batch / lr & 200 / 128 / \(10^{-3}\) \\

\midrule
\multicolumn{3}{l}{\textit{Learned covariate model \(G_\theta\) (DDPM)}} \\
& Training pairs & \(4{,}096\) \\
& Epochs / batch / lr & 200 / 128 / \(10^{-3}\) \\
& Hidden dimension / diffusion steps & 512 / 100 \\

\midrule
\multicolumn{3}{l}{\textit{Budget-DQN and Factorized RL}} \\
& Training episodes & 500 \\
& Replay-buffer capacity & \(10{,}000\) \\
& Batch / hidden / lr & 32 / 64 / \(10^{-3}\) \\
& Training discount factor & Matched to evaluation \\
& \(\varepsilon\) schedule & \(0.20 \rightarrow 0.05\) \\

\midrule
\multicolumn{3}{l}{\textit{GFP value surrogate}} \\
& Latent dimension \(d\) & 32 \\
& Hidden dimension & 64 \\
& Iterations / batch / lr & 200 / 16 / \(10^{-3}\) \\
& State-pool size & 256 \\
& Random-rollout episodes & 64 \\

\midrule
\multicolumn{3}{l}{\textit{GFP referrer-summary network \(L_\eta\)}} \\
& Training parents & 256 \\
& Recruit samples / parent & 64 \\
& Steps / batch / lr & 200 / 128 / \(10^{-3}\) \\

\midrule
\multicolumn{3}{l}{\textit{IID-Population DP}} \\
& Population sample size & \(1{,}024\) \\

\bottomrule
\end{tabular}
\end{table*}

\paragraph{Learned referral dynamics.}
We train the count model \(q_\psi\) on \(2{,}048\) censored triples
\((\vx_i,k_i,y_i)\). Parent covariates are sampled from the synthetic
covariate pool, allocations are sampled from
\(\{1,\ldots,10\}\), and observed recruitment counts are constructed as
\[
y_i=\min\{k_i,C_i\}.
\]
As in the data-calibrated experiments, \(q_\psi\) is a conditional
Poisson model whose rate is parameterized by a two-hidden-layer MLP
with ReLU activations and a softplus output. It is trained for \(200\)
epochs using the censored likelihood in Equation~\eqref{eq:censored-likelihood}.

We train the recruit-covariate model \(G_\theta\) on \(4{,}096\)
parent--recruit covariate pairs generated from the oracle inheritance
kernel. The model is a conditional DDPM whose noise predictor is
conditioned on the parent covariates, the noised recruit covariates, and
the diffusion timestep. It is trained for \(200\) epochs with \(100\)
diffusion steps. Because the covariates are categorical, we convert each
continuous model output into a valid one-hot vector by taking the
argmax within each categorical field.

Since the covariates are discrete, we recover offspring categories from the continuous samples by rounding, following prior work~\cite{kangaslahti2026policy}.

\paragraph{Reinforcement learning baselines.}
We train Budget-DQN and Factorized RL for \(500\) episodes each in the
synthetic planning environment. Both methods use a replay buffer of
capacity \(10{,}000\), batch size \(32\), hidden dimension \(64\), and
learning rate \(10^{-3}\). The exploration rate decreases from
\(0.20\) to \(0.05\). For each experimental setting, the discount
factor used during training is matched to the evaluation value of
\(\gamma\).

\paragraph{IID-Population DP.}
We estimate its marginal population model using \(1{,}024\) covariate
samples from the synthetic planning pool and perform backward induction
over remaining budget and frontier size.

\paragraph{GFP training.}
We use the same GFP architecture as in the data-calibrated
environments. The frontier-value surrogate has latent dimension
\(d=32\) and hidden dimension \(64\). It is trained by fitted value
iteration for \(200\) iterations with batch size \(16\) and learning
rate \(10^{-3}\), using a pool of \(256\) states collected from \(64\)
random-rollout episodes.

The referrer-summary network \(L_\eta\) is a two-hidden-layer MLP with
hidden dimension \(64\) and output dimension \(d\). We train it for
\(200\) gradient steps on \(256\) parent covariates, using \(64\)
recruit samples per parent, batch size \(128\), and learning rate
\(10^{-3}\). During fitted value iteration, \(L_\eta\) is periodically
refreshed using the current value representation \(h_\phi\) and samples
from the fixed learned covariate model \(G_\theta\). We refresh \(L_\eta\) every \(25\) fitted-value-iteration updates.

\section{Computing Infrastructure}
Experiments were conducted with 4× NVIDIA H200 GPUs (with ~1.5 TB host memory and 64 CPU cores).

\end{document}